%% file: main.tex
\documentclass[preprint,12pt]{elsarticle}
\graphicspath{{./}}
\DeclareGraphicsExtensions{.pdf,.jpeg,.png,.jpg,.tiff}
\usepackage{enumitem}
\usepackage{amssymb}
\usepackage{xcolor}

\newlist{todolist}{itemize}{2}
\setlist[todolist]{label=$\square$}
\usepackage{pifont}
\newcommand{\done}{\rlap{\raisebox{0.3ex}{\hspace{0.4ex}\tiny \ding{52}}}$\square$}
\newcommand{\wontfix}{\rlap{\raisebox{0.3ex}{\hspace{0.4ex}\scriptsize \ding{56}}}$\square$}

\usepackage{threeparttable}

\usepackage[caption=false]{subfig}
\usepackage{etoolbox}
\usepackage{siunitx}
\usepackage{color,soul}

\def\citep{\cite}
\usepackage{hyperref}
\hypersetup{
  colorlinks,
  citecolor=blue,
}
\usepackage{multirow}
\usepackage{booktabs}

\journal{Swarm and Evolutionary Computation}

\begin{document}

\begin{frontmatter}


\title{A Prescription of Methodological Guidelines for Comparing Bio-inspired Optimization Algorithms}

\author[upm]{Antonio LaTorre\corref{cor}}
\ead{a.latorre@upm.es}
\author[ugr]{Daniel Molina\corref{cor}}
\ead{dmolina@decsai.ugr.es}
\author[tecnalia]{Eneko Osaba}
\ead{eneko.osaba@tecnalia.com}
\author[ugr]{Javier Poyatos}
\ead{jpoyatosamador@ugr.es}
\author[tecnalia,upv]{\\Javier Del Ser}
\ead{javier.delser@tecnalia.com}
\author[ugr]{Francisco Herrera}
\ead{herrera@decsai.ugr.es}

\address[upm]{Center for Computational Simulation, Universidad Polit{\'e}cnica de Madrid, Spain}
\address[ugr]{DaSCI: Andalusian Institute of Data Science and Computational Intelligence, University of Granada, Spain}
\address[tecnalia]{TECNALIA, Basque Research and Technology Alliance (BRTA), Spain}
\address[upv]{University of the Basque Country (UPV/EHU), Spain}

\cortext[cor]{Corresponding authors.}

\begin{abstract}
Bio-inspired optimization (including Evolutionary Computation and Swarm Intelligence) is a growing research topic with many competitive bio-inspired algorithms being proposed every year. In such an active area, preparing a successful proposal of a new bio-inspired algorithm is not an easy task. Given the maturity of this research field, proposing a new optimization technique with innovative elements is no longer enough. Apart from the novelty, results reported by the authors should be proven to achieve a significant advance over previous outcomes from the state of the art. Unfortunately, not all new proposals deal with this requirement properly. Some of them fail to select appropriate benchmarks or reference algorithms to compare with. In other cases, the validation process carried out is not defined in a principled way (or is even not done at all). Consequently, the significance of the results presented in such studies cannot be guaranteed. In this work we review several recommendations in the literature and propose methodological guidelines to prepare a successful proposal, taking all these issues into account. We expect these guidelines to be useful not only for authors, but also for reviewers and editors along their assessment of new contributions to the field.
\end{abstract}


\end{frontmatter}

\section{Introduction}
\label{sec:intro}

Bio-inspired algorithms in the field of optimization is a mature research area. The number of contributions submitted to conferences and journals of this area increases sharply every year \cite{del2019bio}. However, a major fraction of these proposals do not prove the goodness of new algorithms appropriately. It is often the case that a work presenting a new bio-inspired algorithm raises doubts in regards to the true contribution of the new proposal. As a result, these concerns put at risk its acceptance by the research community or, alternatively, its capacity to assess the true contribution and significance of the proposed research.

There are a number of reasons for this noted fact, ranging from low-quality papers to works lacking originality \cite{molina:20}. In those cases, there is little to do but to continue investigating toward reaching better results. On the other hand, there is a number of works whose contribution appear to be significant, but that can not be accepted for several reasons. These include, but are not limited to, experimental flaws, questionable/insufficient validation efforts, or a weak discussion of the results. Although these practices could be easily avoided, their repeated occurrence makes them a crucial problem: rigorous experimental practices are needed so that the community could embrace the conclusions drawn from a research work, eventually leading to meaningful advances in this research field.

Several papers can be found with suggestions about important issues found in experimental benchmarks and comparison among meta-heuristics. Each of them focuses on a specific aspect, such as how to design the experiments \citep{stanley66_influen_fisher_desig_exper_educat} or how to select and interpret statistical tests to assess the relative differences among algorithms \citep{statistics}. However, to the best of our knowledge, there is no prior work that deals, at the same time, with different relevant issues that could ultimately jeopardize the fairness assumed in the performance comparisons among techniques. When the goal is to discriminate which algorithm performs best among a set of possible choices, fairness should be an unwavering principle. This includes the design of the benchmark, the selection of performance metrics, and the analysis and discussion of the results. Without this principle being guaranteed along the experimental workflow, conclusions extracted from these studies will remain in doubt.

The main objective of this manuscript aligns with the above remarks. Specifically, we review the literature background and provide a set of useful guidelines intended for researchers to avoid common mistakes in experiments with bio-inspired meta-heuristics. Those bad practices could eventually generate doubts about the fairness of the comparisons reported therein. Our methodological approach is pragmatic, mainly aimed at making it easier for new researchers entering this area to prepare an experimental section under high quality standards. We start our study by exploring current approaches for algorithms' analysis. We pay special attention to bad practices, identified not only in this work but also in previous literature. All this information can be found in Section \ref{sec:analysis}.

Then, we propose 4 guidelines that authors introducing a new algorithmic proposal should take into account to boost their chances to get their work embraced by the community. We provide a brief motivation for each of them in the following paragraphs, whereas Sections \ref{sec:guideline1}--\ref{sec:guideline4} present detailed methodological guidelines for the elaboration of successful proposals:

\begin{itemize}[leftmargin=*]
\item \textbf{Guideline \#1: Benchmarks.} Sometimes the benchmark is a real-world problem. In these cases, the benchmark gauges how the proposed algorithm tackles the problem at hand. By contrast, in other cases the proposal is compared against other reference algorithms by using a benchmark specially designed to test their performance. In any case, the selection of the appropriate benchmark is an important issue, since the conclusions that can be extracted from the study depend deeply on the test bed. Unfortunately, the chosen benchmarks frequently present some features that might favor algorithms with a particular bias. This is, of course, not desirable for the sake of fairness in the subsequent comparisons. Thus, the results obtained by the newly proposed solver must be analyzed by taking into account the different characteristics of the test problems covered by the benchmark at hand.

\item \textbf{Guideline \#2: Validation of the results.} The presentation of raw results arranged in full-page tables is, today, not enough. A proper validation of the results from a statistical point of view should always be provided along with the aforementioned tables. In this sense, it is important that not only statistical tests are used, but also that the correct ones are applied. It is quite usual to find parametric tests that are used without ensuring that the assumptions required for those tests are met by the obtained results. In addition, we also recommend visualization techniques for comparative analysis. They can summarize a significant amount of information in a condensed representation that can be quickly grasped and interpreted by the reader. 

\item \textbf{Guideline \#3: Components analysis and parameter tuning of the proposal.} The hypotheses of the proposal must be clearly stated at the beginning of the paper, and discussed once the results have been validated. Moreover, the authors should conduct a thorough analysis of the results considering, at least, the following aspects: search phases identification (balance between exploration and exploitation), components and complexity analysis (individual analysis of the contribution of each of the components of the overall method, and their complexity), parameter tuning of the algorithm, and statistical comparison with state-of-the-art algorithms (as described in Guideline \# 2).

\item \textbf{Guideline \#4: Why is my algorithm useful?} Finally, prospective contributors should clearly state why their proposed algorithm should be considered relevant by the rest of the community. In this guideline we discuss this issue in depth from different points of view. We also suggest several reasons for which a new proposal poses an advance in knowledge (i.e. it is found to be competitive against state-of-the-art methods, it presents methodological contributions that stimulate further research, or other reasons later elaborated).
\end{itemize}

In order to illustrate each of the problems discussed in this contribution, we will resort to different use-cases coming from our previous experience or especially tailored for the purposes of this study. Data utilized for each of these exemplifying problems may vary, as not all situations can be clearly explained with one single example. We also provide a case study in Section \ref{sec:case_study} that describes the process of designing and evaluating new algorithms  according to the methodology proposed herein. It embraces all the methodological guidelines by, first, selecting a standard benchmark (CEC'2013 LSGO), a performance measure (ranking) and the reference algorithms. Then, we conduct a proper statistical validation of the results compared with those of the reference algorithms. We also use visualization techniques to offer a more clear view of the results. To continue, the contribution of each component of the new algorithm is analyzed to ensure that all of them contribute to the results of the overall method. The case study finishes with a discussion on the usefulness of the new proposal. As can be seen, it properly covers all the methodological guidelines proposed in this contribution. 

As a summary, the main key elements of this work are:
\begin{enumerate}[leftmargin=*]
    \item A literature review, with an emphasis on the identification of bad practices in the analysis of new algorithmic proposals.
    \item Four methodological guidelines to help authors achieve contributions adopted by the community.
    \item A case study, as described in the previous paragraph, that simulates the process of proposing a new algorithm by following the aforementioned four methodological guidelines.
\end{enumerate}

The remainder of this paper is organized as follows. Section \ref{sec:analysis} discusses several previous useful guidelines and recommendations in the literature. Sections \ref{sec:guideline1} through \ref{sec:guideline4} present and discuss the guidelines proposed in this work, whereas in Section \ref{sec:case_study} we provide several case studies covering some of these guidelines. Finally, Section \ref{sec:conclusions} concludes the study.

\section{Relevant issues for the proposal of methodological guidelines}\label{sec:analysis}

In any field of science, it is crucial to work under correct and unbiased experimental conditions, and to conduct a rigorous and adequate analysis of the obtained results. However, sometimes there are small aspects that can lead to inadvertently biased comparisons, partially benefiting a certain type of algorithms over the others.

In this section we review prior work in the literature, advising constructively against issues that could generate objective doubts about the strength of the experimental claims. Specifically, we revisit several particular topics of relevance for the current study: an inadequate or incomplete description of an algorithm (Section \ref{sec:alg_desc}), the presence of bias in the search process (Section \ref{sec:lit_benchmark}), relevant features that should be taken into account when selecting benchmarks (Section \ref{sec:features}), prior studies focused on the validation of experimental results (Section \ref{sec:lit_results}), and existing works on component analysis and parameters tuning (Section \ref{sec:lit_analysis}). Topics tackled in this first background analysis have a straightforward connection with our methodological guidelines given in Sections \ref{sec:guideline1} to \ref{sec:guideline4}.

\subsection{Inappropriate description of the algorithm}
\label{sec:alg_desc}

This is a common issue in many proposals, especially in those of more advanced methods. Reproducibility of scientific results should always be required and this is not possible if some implementation details are missing \cite{Johnson2002,Hellwig2019}. This includes not only a high-level description of the algorithm, but also implementation details, dependencies, parameters values, etc. Furthermore, if the proposal is based on a pre-existing method, differences with regards to the base algorithm should be stressed. Finally, and although this is still not mandatory in most journals and conferences, we encourage authors to freely distribute the source code of their algorithms, to allow other users to better replicate their results.

\subsection{Bias in the search process}
\label{sec:lit_benchmark}

One of the most critical decisions when evaluating an algorithmic proposal is the selection of the benchmark used to show its goodness.
Unfortunately, for many papers the testbed was proposed by the same authors, and is normally a combination of well-known synthetic theoretical functions. Moreover, the only measure of performance is often the benchmark proposed. The design of a good benchmark is not an easy task, and they can be used for benefiting newly proposed methods by exploiting any bias in the search algorithm \citep{weise12_evolut_optim}.
\begin{itemize}[leftmargin=*]
\item \emph{Optima close to the center of the domain search:} one of the most typical sources of bias is the tendency of some algorithms to explore with more intensity in the surroundings of the center of the domain search. This issue was first discussed and termed \textit{structural bias} in \citep{struct_bias}. Many versions of Genetic Algorithms (GA), Particle Swarm Optimization (PSO), or Differential Evolution (DE) \citep{Storn1997}, have exploited this characteristic, as it has been traditionally where the optimum of the problem under analysis was located. Those algorithms, on the other hand, tend to exhibit a bad performance near to the bounds of the domain search \citep{hue16_bounds}. In \cite{piotrowski16} and \cite{Caraffini2019} a detailed experimentation about the structural bias in search algorithms is given. Avoiding this kind of bias in the design of algorithms is not easy, but at least they should not be evaluated on benchmarks favored by these biases during exploration. One popular approach to avoid having the optimum into the center of the domain search is shifting.

\item \emph{Sensitivity to the coordinate system:} another possible source of bias emerges when the exploration of the search domain is done mainly by moving along the directions of the coordinate system. In this regard, some algorithms have proven to be very sensitive to the coordinate system \citep{Price2017}. Some benchmark functions are rotated to test the invariance of the algorithms to such transformations. Ideally, the algorithm should be invariant to these rotations, such as Covariance Matrix Adaptation Evolution Strategy (CMA-ES) \cite{Hansen2001}) and Black-Box DE \citep{Price2017}.
\end{itemize}

In order to compare algorithms designed to avoid these sources of bias, several benchmarks taking into account these issues have been proposed. This allows a fairer comparison between algorithmic proposals, that can be compared on the same testbed. In particular, in real-parameter optimization, several benchmarks have been proposed since 2005 to date \cite{funcec2005,funcec2011real,funcec2013real,funcec2014real,funcec2017real}. All these benchmarks try to avoid the first source of bias by shifting the location of the global optimum. Furthermore, in the more recent testbeds an increasing number of functions have been rotated, featuring more complex landscapes. A more detailed view of the different benchmarks proposed and their evolution can be found in \cite{Molina2018a}.

On a closing note in this matter, benchmarks should definitely evaluate the sensitiveness of the algorithms to different sources of bias as the ones mentioned previously. However, when moving from the academic realm towards real-world scenarios, it is important to bear in mind that the goal is not to find a good \emph{meta-heuristic} approach, but rather to solve a given problem efficiently. This being said, the availability of a priori information on any bias of the problem to be solved should be exploited. Indeed, if a solver exploiting a certain source of bias known to exist in the problem, then this solver should be preferred.  Actually, benchmarks are useful to identify algorithms with a good performance on problems with similar sources of bias. These thoughts concur with recent work around the exploitation of problem-specific knowledge when designing an optimization algorithm aimed to solve it efficiently (grey-box optimization, see e.g. \cite{wu2017using} and references therein). In summary, if the objective of the study is to solve a given problem, taking advantage of possible sources of bias when designing the algorithm is convenient and advisable.

\subsection{Relevant features of benchmarks} \label{sec:features}

Different properties related to the landscape of the functions composing the benchmark should be addressed for a fair analysis of the behavior of the proposed solver(s). Among them:

\begin{itemize}[leftmargin=*]
\item \textbf{Separability of the components:} some functions can be easily solved by optimizing each dimension individually, so it is crucial not to use only separable functions in the benchmark. This is the case, for example, of the CEC'2008 LSGO benchmark proposed in \cite{Tang2007}, in which many functions are of this type. In more recent benchmarks, especially in the field of large-scale optimization, the focus is on evaluating the capability of the algorithms to identify existing subcomponents in the functions. If the new proposal deals with this kind of problem, it should be tested on a benchmark that allows evaluating this characteristic, such as e.g. the one proposed in \cite{Li2013a}.
  
\item \textbf{Dimensionality of the problems:} another important issue is the dimensions of the benchmark, because some algorithms are designed to work properly only for very low-dimensional problems. As the size of the search domain increases exponentially with the dimensionality of the problem, the so-called \textit{curse of dimensionality} \citep{bellman1966dynamic} poses a significant computational challenge. This is particularly the case for well-known algorithms such as PSO \citep{vandenbergh04_cooper_approac_to_partic_swarm_optim} and DE \citep{neri09_recen_advan_differ_evolut}. Actually, as the performance of most algorithms degrades when the dimension grows, the current trend is to develop specific algorithms for problems with higher dimensionality. Nonetheless, it is increasingly important to offer robust performance for a medium range of dimension values. Some benchmarks are designed to evaluate the performance of algorithms in problems of small to moderate size \cite{funcec2005,funcec2011real,funcec2013real,funcec2014real,funcec2017real}, whereas others aim at problems of a much larger size \cite{Tang2007,Tang2010,Li2013a}. These are problems of a very different nature and, normally, algorithms with an outstanding performance over one type of problems do not perform as such when applied to other types of problems. For example, strategies such as computing the covariance matrix of solutions as done by \cite{Hansen2001} do not scale up nicely for problems of larger size. Furthermore, these latter problems require increased exploration abilities of the algorithm to cover a much broader search space, which has a significant cost in terms of fitness evaluations.

\item \textbf{Number of optima of the problems:} There are functions that have only one optimum, and other that have several optima called multi-modal functions. Multi-modal functions can have several global optima, with the same fitness value, and also multiple local optima, with different (worse) fitness values. The presence of local optima increases the difficulty of the optimization process because the algorithm can be stuck in them. This is the case of algorithms with a strong elitist behavior \cite{olivetoHowEscapeLocal2018}. The difficulty increase is due to the fact that the black-box optimization algorithms are not aware of the number of local optima, although several works have proposed methods to estimate their number \cite{hernandoEvaluationMethodsEstimating2013a}.

\item \textbf{High-conditioning problems:} A high-conditioning or ill-conditioning function is one in which a small change in the variables of the solution implies a large change in its fitness value. This means that the correct solution/answer to the problem becomes hard to find for optimization algorithms \cite{suttonDifferentialEvolutionNonseparability, bagheriSolvingOptimizationProblems2019}.

\item \textbf{Noisy functions:} finally, noise is another important factor that has not been widely considered in the literature. However, this is changing recently in several recent benchmarks that also consider this issue. These benchmarks, such as the BBOB \cite{Finck2010} or the Nevergrad \cite{nevergrad} benchmarks, explicitly include functions with different degrees of noise that resemble real-world scenarios in which noise can be a very important issue. Despite this recent interest, very few studies have hitherto dealt with noisy functions \citep{lit_noisy,Jin2005}.
\end{itemize}

\subsection{Validation of the results}
\label{sec:lit_results}

Selecting competitive algorithms to be included in the comparison is another crucial aspect in benchmarking. In the current literature many different algorithms can be found and chosen to be reference algorithms for a given benchmark. Unfortunately, there is no clear criterion to make such a selection. Although good practices usually suggest comparing against the most recent state-of-the-art algorithms, it is often the case that authors only compare their proposal against basic or very similar versions of other algorithms, as was spotted in \cite{garcia-martinez17_since_cec_compet_real_param_optim}.

In this context, studies comparing different algorithms are scarce, and the results reported therein strongly depend on the problem(s). In \cite{civicioglu11_concep_compar_cuckoo_searc_partic} a benchmark of classic functions was used to compare among Cuckoo Search (CS), PSO, DE, and Artificial Bee Colony (ABC). The study concludes that the best results were obtained by CS and DE, and the worst ones were those rendered by ABC. On the other hand, for a different problem \citep{osuna-enciso13_compar_natur_inspir_algor_multi}, ABC was found to perform best, followed by DE and PSO.

For the methodological part of the comparisons, there are far more studies. Statistical tests, for instance, lay at the core of prior contributions on this matter. However, such contributions are frequently written from a statistical point of view -- like the one by Demšar \citep{demsar} -- making it difficult for researchers in this field to embrace their methodological recommendations. More recently, some tutorials have tried to bring together the fields of meta-heuristics and inferential statistics \citep{statistics}. Some examples can be found in \cite{Whitacre_useof}, in which a statistical treatment is suggested for distinguishing between measurements of performance in adaptive evolutionary algorithms. Another good example is \citep{garcia08_study_use_non_param_tests}, which shows that in a popular real-parameter benchmark (CEC'2005), conditions needed for running parametric hypothesis tests did not hold, and non-parametric tests were thus recommended. More recently, in \cite{garcia10_tests}, some recommendations for the comparison of evolutionary algorithms are provided, which can be even extrapolated to machine learning benchmarks.

Another important issue from a methodological point of view is the assessment of the performance of bio-inspired algorithms from the perspective of the experimental design. Some studies \cite{Johnson2002} provide general recommendations to design experiments for the comparison of algorithms in a similar way to what we do in this contribution. However, these recommendations are far more general as it targets a broader scope (the design of algorithms and not bio-inspired optimization methods, specifically). This difference in the target of the proposed guidelines makes it miss some important issues inherent to bio-inspired optimization methods that we cover in this contribution. Other works, although focused on optimization methods, are more specific in their recommendations, targeting specific issues such as the selection of problems and performance measures \cite{McGeoch2002,Eiben2002,Halim2020}. Finally, other studies are more oriented to the analysis and definition of experimental frameworks such as those used in CEC special sessions and the COCO framework \cite{Hellwig2019}. While this is also a very relevant contribution to the field of experimental design in the context of evolutionary algorithms, it deals with the problem from a different perspective and should be considered as a complement to the guidelines here presented.

\subsection{Components analysis and parameter tuning}
\label{sec:lit_analysis}

New proposals are frequently based on previously existing algorithms to which certain components have been added/replaced. However, the addition/replacement of new components is not always adequately justified. This is a fundamental design flaw that can contribute to make more and more complex algorithms in which the new additions only report a marginal contribution to the overall performance of the method. Authors should clearly discuss about the contribution of each new component in order for the new proposal to be considered significant.

Another important research topic in the design of meta-heuristic algorithms is the selection of the values of their parameters. Indeed, parameters can be a double-edged sword. On the one hand, they grant flexibility to control the search behavior of the algorithm. On the other hand, finding the parameter values that lead to the best search performance is another optimization problem itself \citep{eiben99_param_contr_evolut_algor}. For this reason, there is a long literature record of studies dealing with the best parameter values for different meta-heuristic algorithms, such as GA \cite{grefenstette86_optim_contr_param_genet_algor}, PSO \cite{trelea03_partic_swarm_optim_algor}, or DE \cite{qin09_differ_evolut_algor_with_strat,neri09_recen_advan_differ_evolut}.

The tuning of parameters can be carried out by means of different automatic tuning tools. There are several consolidated tools of this nature, with different features \citep{tuning_method}. F-RACE \cite{frace} and I-RACE \cite{irace} are iterative models, which, at every step, evaluate a set of candidate parameter configurations, discarding several of them along the search. These methods remove candidates upon the result of statistical comparisons, e.g. two-way analysis of variance. I-RACE is an implementation of Iterative F-RACE that includes several extensions, such as a restart using normal distributions. REVAC \cite{REVAC}, on the contrary, relies on an Estimation of Distribution Algorithm (EDA). For each parameter, REVAC starts by sampling an uniform distribution of values. Then, at each step it reduces the value range of each parameter by using specially designed transformation operators, considering an entropy measure. ParamILS \cite{ParamILS} is an iterative local search algorithm that, from a default parameter configuration, applies a local search with random perturbations to improve the configurations.

\section{Guideline \#1: Benchmarks} \label{sec:guideline1}

The first decision that a researcher must face when preparing a new contribution in the field of optimization is the selection of the benchmark to test the newly proposed algorithm(s). Once this is done, the authors must identify relevant algorithms to compare the obtained results and guarantee the significance of conclusions drawn therefrom. This first guideline deals with these two crucial aspects: the selection of the benchmark (Section \ref{sec:guideline1_benchmark}), and the reference algorithm(s) to which the proposal is compared (Section \ref{sec:guideline1_ref_algs}).

\subsection{Selection of the benchmark}
\label{sec:guideline1_benchmark}

As mentioned before, this first decision is one of the most important factors to prove the quality and performance of an algorithm. In real-world problems, this is not a decision at all, because the benchmark is the problem to tackle. In contrast, when designing and improving meta-heuristic techniques, the selection of a benchmark is an important decision to take. This issue is common for all types of optimization. During the last years, benchmarks have been proposed for several types of optimization (such as combinatorial and numerical optimization) with the main goal of becoming a standard for future comparisons. Without loss of generality, during the following section we will focus on numerical optimization, yet all conclusions and recommendations given hereafter are applicable regardless of the domain. 

In the field of numerical optimization, special sessions devoted to benchmarking have taken place in reference events such as the IEEE Congress on Evolutionary Computation or the Genetic and Evolutionary Computation Conference. In these events, participants could compare their algorithms in a controlled environment with a homogeneous set of functions \cite{Molina2018a, nikolaus_hansen_2019_2594848}. Nonetheless, the efforts towards providing standard benchmarks and tools for the comparison of bio-inspired optimization methods is not limited to special sessions and competitions, as shown below:
\begin{itemize}[leftmargin=*]
    \item Special Sessions and Workshops:
    \begin{itemize}[leftmargin=*]
        \item IEEE Real-Coding Special Session (since 2005)
        \item Black-Box Optimization Benchmarking (BBOB) Workshops at GECCO and PPSN (since 2009)
        \item Black-Box Optimization Competition (BBComp) (since 2015)
        \item IEEE Large-scale Global Optimization Special Session (since 2008)
        \item Benchmarking Workshop at GECCO and PPSN (since 2020)
        \item Other (more specific) special sessions: constrained, multimodal, real-world, etc.
    \end{itemize}

    \item Tools:
    \begin{itemize}[leftmargin=*]
        \item Comparing Continuous Optimizers (COCO) framework\footnote{\url{https://github.com/ttusar/coco/}}
        \item IOHProfiler\footnote{\url{https://iohprofiler.github.io/}}
        \item Nevergrad\footnote{\url{https://github.com/facebookresearch/nevergrad}}
        \item TACO: Toolkit for Automatic Comparison of Optimizers\footnote{\url{https://tacolab.org/}}
        \item P. N. Suganthan's CEC benchmarks\footnote{\url{https://github.com/P-N-Suganthan}}
        \item IEEE CEC LSGO Benchmark\footnote{\url{http://www.tflsgo.org/special_sessions/cec2019.html\#original-code}}
    \end{itemize}

    \item Other initiatives:
    \begin{itemize}[leftmargin=*]
        \item Benchmarking Network\footnote{\url{https://sites.google.com/view/benchmarking-network/}}
        \item Cost Action CA15140 (ImAppNio)\footnote{\url{https://imappnio.dcs.aber.ac.uk/}}
    \end{itemize}
\end{itemize}

However, some works simply overlook these standard benchmarks and tools. Instead, they rather choose their own subset of functions to evaluate their proposal. This is problematic for several reasons. First, as follows from Section \ref{sec:lit_benchmark}, it is very hard to know whether there is any bias in the selection of the functions from the point of view of the performance of the algorithm under consideration. Secondly, many different benchmark functions exist (or can be defined). Therefore, it becomes very difficult to appraise the different characteristics of all such benchmark functions. Finally, comparisons with other reference methods usually imply running them by the same authors of the new method, as it is very unlikely that multiple studies would have focused on the same selected functions or experimental conditions. As a result, assessing the quality of a new contribution that does not use standard benchmarks gets almost impossible to accomplish, and therefore should not be given credit by the community.

However, there are two special situations in which researchers have no other option but to use \textit{ad-hoc} generated problem instances: i) when the problem to be solved has never been tackled in the previous literature, and hence no benchmark can be found; or ii) when a real-world problem is under consideration, with specific requirements and constraints. In these cases, the instance generation process must be deeply detailed, and all the instances generated should be shared for other researchers to replicate and improve upon the presented results. Furthermore, for any of these two alternatives, practitioners should generate a benchmark as realistic and general as possible. 

On the other hand, one should also be very careful when selecting a benchmark for the experimentation to be carried out. Benchmarks in the literature have been conceived with some objectives in mind, and are appropriate to test certain characteristics of the algorithms under evaluation. A non-exhaustive list of these characteristics follows:

\begin{itemize}[leftmargin=*]
\item \textbf{Bias avoidance of the search algorithm:} In order to avoid the problems described in Section \ref{sec:lit_benchmark}, it is highly advisable that the optimum is not located at the center of the domain search (e.g. by shifting). Furthermore, rotation should be enforced to test the sensibility of the algorithm to the coordinate system.
    
\item \textbf{Sensitivity to the number of local optima:} The number of local optima of a function is another important characteristic of a test problem. Unless properly considered in the algorithmic design, search spaces with multiple local optima may negatively affect the convergence of metaheuristics towards the global optimum. In this kind of problems, multiple local optima can act as basins of attraction, preventing the algorithm from reaching the global optima if it does not correctly balance its exploration/exploitation ratio.
\end{itemize}

Additionally, benchmarks normally establish the experimental conditions under which algorithms should be evaluated. In particular, it is very frequent that a benchmark selects a common:

\begin{itemize}[leftmargin=*]
\item \textbf{Measure of performance:} In evolutionary algorithms it is usual to use the mean error obtained for the different runs. However, there are more adequate measures and indicators of performance for dynamic optimization \cite{helbig13_perfor_measur_dynam_multi_objec_optim_algor}, and multi-objective optimization \cite{mirjalili15_novel_perfor_metric_robus_multi}. Additionally, other performance indicators, such as running time of memory footprint, could also provide interesting insights on the behavior of the algorithms (especially in real-world scenarios). Nonetheless, this should be carefully considered as those measures might be biased depending on external factors such as the programming language of choice and not the algorithm itself, which can hinder the comparison.

\item \textbf{Stopping criterion:} In order for a comparison of several algorithms to be fair, all of them must conduct a similar effort in finding a solution. This is normally achieved by establishing a common stopping criterion. If this were not the case, one algorithm could be stopping when a predefined precision is reached whereas other algorithm could run until a maximum budget of fitness evaluations is exhausted. The results of both algorithms are not comparable and this is why most benchmarks define a common stopping criterion. It also allows grasping the full picture over the performance of algorithms, as different methods may yield different convergence rates, and the results of the comparison might differ depending on the checkpoint at which they are evaluated. This issue will arise and will be discussed in the use cases presented in Section \ref{sec:case_study}.
\end{itemize}

Finally, there is another relevant characteristic from a design perspective that is not specifically linked to the functions themselves, but to the algorithms that solve those functions. The algorithms use the fitness function to guide the search process, but sometimes only the ranking of the solutions computed from fitness values is actually used. In this sense, some recent algorithms such as the Firefly algorithm \citep{FA} or the Grasshopper optimization algorithm \citep{GOA} use the quantitative
information provided by the fitness function to guide the search process, whereas others propose mutation and/or selection methods \cite{DERank1} that only require to know whether one solution is better than the other. In some real-world scenarios, the last approach can be very beneficial as it simplifies the process of defining the fitness function.

\subsection{Selection of the reference algorithms} \label{sec:guideline1_ref_algs}

Another important issue, which is actually related to the previous guideline, is the selection of the reference algorithms to include in the comparison. On the one hand, if the proposed algorithm relies on some other basic algorithms, these should be included in the comparison to check the individual contribution of each of them, as we will discuss in detail in Section \ref{sec:guideline3}. On the second hand, once the benchmark has been selected, the best-so-far methods for that particular benchmark should be also considered in the comparison. Unfortunately, many papers fail to compare their proposed algorithm against competitive counterparts \citep{garcia-martinez17_since_cec_compet_real_param_optim}. 

A well-informed experimentation should, at least, include the best algorithms in the special session where the benchmark was originally proposed (as it is usually the case). We refer to \cite{Molina2018a} for an updated review on special sessions and competitions on continuous optimization.

Finally, authors should also consider similar algorithms, not only from the same \emph{family} (e.g., PSO, DE, GA...) but also the base algorithm, if the proposal is an improvement over a previous algorithm, or other similar approaches (for example, different memetic algorithms with a common local search). Our claim in this regard is to stop comparing new methods with classic algorithms that have been clearly outperformed by newer ones. Comparisons adopting this misleading strategy should be avoided for the questionable scientific contribution of their proposal.

\section{Guideline \#2: validation of the Results}
\label{sec:guideline2}

Just as important as a correct experimentation design (see Guideline \#1) is a principled validation procedure for the benchmark. For this purpose, we emphasize on two different tools: statistical analysis and comparative visual analysis. Both approaches are covered in the following two subsections: statistical analysis (Section \ref{sec:guideline2_stat}), and visualization techniques for comparing meta-heuristics (Section \ref{sec:guidelines2_visualization}).

\subsection{Statistical analysis: non-parametric tests and beyond} \label{sec:guideline2_stat}

Statistical comparison of results should be deemed mandatory in current benchmarks among bio-inspired algorithms. However, even if statistical comparisons are made in studies reported nowadays, they are not always carried out properly. In inferential hypothesis testing, there are some popular methods, such as the t-test or the ANOVA family of tests. However, these tests are referred to as \emph{parametric tests} because they assume a series of hypotheses on the data on which they are applied (i.e., on the parameters of the underlying distribution of the data). If such assumptions do not hold (for example, the normality assumption for the results), the reliability of the tests is not guaranteed, and alternative approaches should be considered. Thus, either these conditions are checked to be true (i.e., by using a normality test to prove the normality in the distribution), or another type of tests that do not make these assumptions should be used instead. This is the case of \emph{non-parametric} tests, which do not assume particular characteristics for the underlying data distribution. This non-parametric nature must be seen as an advantage over the aforementioned parametric tests (for their independence with respect to data) but also as a limitation (non-parametric tests are less powerful) imposed by the nature of the underlying data, that do not satisfy the requirements to use the more powerful parametric methods.

As a result of this, parametric tests should be preferred whenever they can be safely used (i.e. whenever the hypotheses on the underlying data distribution are met). Unfortunately, this often fails to be the case when comparing the results of bio-inspired algorithms. Consequently, non-parametric tests should be used instead \citep{statistics}. A common error (less frequent in current research) is to apply parametric tests without checking if the required hypotheses are satisfied.

In the following paragraph we provide a workflow to decide which kind of test to choose. It consists of the following steps:

\begin{enumerate}[leftmargin=*]
	\item Check the conditions required for the application of the parametric test of choice (normally, Student's t-test).
	\begin{enumerate}
  		\item Normality: Shapiro-Wilk test \cite{Shapiro1965} or Kolmogorov-Smirnov. Shapiro-Wilk should be used with smaller sample sizes \cite{KS}.
		\item Homocedasticity (equal variances): Levene's test \cite{Levene1960}.
	\end{enumerate}
	\item If both conditions are satisfied, apply Student's t-test \cite{Sheskin2007}.
	\item If only normality can be guaranteed, then Welch's t-test alternative is considered \cite{Welch1947}.
	\item If none of the assumptions on the underlying distribution holds, then the non-parametric Wilcoxon signed-rank test is used \cite{Wilcoxon1945}.
\end{enumerate}

\begin{table}[!htbp]
\caption{Recommended tests according to data characteristics}
\label{table:statistical_tests}
\centering
\resizebox{\columnwidth}{!}{\begin{tabular}{lcc}
\toprule
\multicolumn{1}{l}{Conditions}     & Equal variances         & Unequal variances     \\ \midrule
\multicolumn{1}{l}{Normally distributed}     & Paired Student's t-test & Paired Welch's t-test \\ \midrule
\multicolumn{1}{l}{Not normally distributed} & \multicolumn{2}{c}{Wilcoxon signed-rank test}       \\ \bottomrule
\end{tabular}}
\end{table}

Once the appropriate test has been selected, the comparison can be carried out. First, the ranking of each algorithm over the whole benchmark must be computed, and the significance of the differences in the ranking values must be tested. Friedman rank-sum test can serve for this purpose \cite{Daniel1990}. If differences are declared to be statistically significant by the Friedman test, then we proceed to the pairwise comparisons with the test selected from Table \ref{table:statistical_tests}. In those pairwise comparisons a reference algorithm is compared against all the other methods selected for validation. Normally, the reference algorithm is selected to be the one with the best average ranking or, alternatively, the new proposal presented in the work.

Another typical oversight noted in the literature is to neglect the accumulated error. A statistical test for two samples, like the Wilcoxon's test, has an estimated error, but this error increases with each pair of comparisons. Thus, when simultaneously comparing the results of our proposal with those attained by several other algorithms, the application of Wilcoxon's test (or others such as the t-test) is totally discouraged, because it cannot ensure that the proposal is statistically better than all the other reference algorithms. Thus, once the pairwise p-values have been computed, a correction method must be used to counteract the effect of multiple comparisons, by controlling either the family-wise error rate, or the false discovery error rate \cite{Aickin1996}. Several procedures have been proposed to this end, among which Bonferroni-Dunn \cite{Dunn1961}, Holm-Bonferroni \cite{Holm1979}, Hochberg \cite{Hochberg1995} and Hommel \cite{Hommel1988} are the most widely used \cite{statistics}.

Also linked to statistical validation, another recommendation is to provide the p-values of tests carried out in the experimentation. However, we note that p-value, as such, is not a totally objective measure, as it is highly dependent on the sample size \cite{lin13_resear_commen_too_big_to_fail}. Section \ref{sec:case_study} presents several examples of comparisons that goes through the methodological steps prescribed in this second guideline.

Traditionally, the standard null hypothesis testing methods we just described have been used for comparing the performance of different metaheuristic algorithms. This has produced over the years a large number of different post-hoc tests and graphical representations that ease the process of evaluating which algorithm performs best on average, with statistical significance (e.g. critical distance plots). However, much criticism has arisen lately in different aspects of these tests that suggest that a step further should be made towards alternative means to assess the statistical relevance in performance comparison studies. To begin with, the use of the so-called significance parameter (often denoted as $\alpha$) clashes with its lack of interpretability, and does not link directly to the performance differences observed among the counterparts in the benchmark. Furthermore, conclusions drawn from non-statistical hypothesis testing are largely sensitive with respect to the number of samples used for their computation, as well as the number of algorithms and problems over which the study is made. The work by Benavoli et al in \cite{benavoli2017time} lit a light on this matter, and proposed the use of Bayesian analysis for comparison analysis. The Bayesian paradigm makes statements about the distribution of the difference between the two algorithms under comparison, which can be of help when the null hypothesis significance test (NHST) does not find significant differences between them. The rNPBST package \cite{carrasco2017rnpbst} and the jMetalPy framework \cite{BENITEZHIDALGO2019100598} are useful tools to apply these tests. 

The use of different tests can help putting the results in context. As it is mentioned in \cite{carrasco2020recent}, authors encourage the joint use of non-parametric and Bayesian tests in order to obtain a complete perspective of the comparison of the algorithms’ results: \emph{``While non-parametric tests can provide significant results when there is a difference between the compared algorithms, in some circumstances these tests do not provide any valuable information and Bayesian tests can help to elucidate the real difference between them''} \cite{carrasco2020recent}. Practitioners must consider this possibility to complement well-known non-parametric tests when they do not provide a full difference among algorithms. 

\subsection{Visualization techniques for comparative analysis} \label{sec:guidelines2_visualization}

Visualization techniques are other useful methods to report results when comparing several bio-inspired algorithms. The main advantage of these approaches over reporting raw data in tables is that they can be much easily interpreted by the reader. They have also the ability to summarize the information covered by one or even multiple tables.

\begin{figure*}[!ht]
  \centering
  \subfloat[\label{fig:multimodal_chart}]{\includegraphics[width=0.5\linewidth]{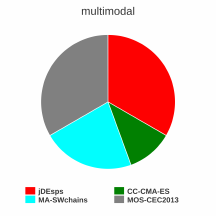}}
  \subfloat[\label{fig:shifted_chart}]{\includegraphics[width=0.5\linewidth]{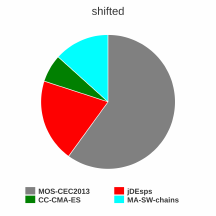}}\\
  \subfloat[\label{fig:all_funcs_chart}]{\includegraphics[width=0.5\linewidth]{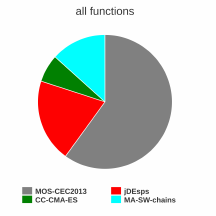}}
    \subfloat[\label{fig:radar_chart}]{\includegraphics[width=0.5\linewidth]{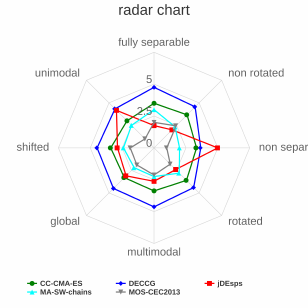}}
    \caption{Different visualizations for the comparison of the performance of several algorithms on the CEC'2013 LSGO benchmark: (a) Average ranking of algorithms on different types of functions; (b) Fraction of functions for which a specific algorithm obtained the best results; (c) Fraction of multimodal functions for which a specific algorithm obtained the best results; (d) Fraction of shifted functions for which a specific algorithm obtained the best results.}
    \label{fig:charts}
\end{figure*}

In Figure \ref{fig:charts} we provide an example of some visualizations that illustrate the performance of several algorithms on the CEC'2013 LSGO benchmark. Figure \subref{fig:radar_chart} uses a radar chart to visualize the average ranking of each bio-inspired algorithm on different groups of functions. Each group has been defined according to some common characteristics present in many state-of-the-art benchmarks: degree of separability, modality, etc. Figures \subref{fig:all_funcs_chart}-\subref{fig:shifted_chart} provide an alternative view on the same data: it does not depict their average behavior, but instead the number of times in which one algorithm obtained the best overall results for problems belonging to each of the previously defined categories. In Figure \subref{fig:all_funcs_chart}, the whole benchmark is considered, whereas Figures \subref{fig:multimodal_chart} and \subref{fig:shifted_chart} show the results for multimodal and shifted functions, respectively.

\begin{figure}[!ht]
    \centering
    \subfloat[\label{fig:conv_11}]{\includegraphics[width=0.7\linewidth]{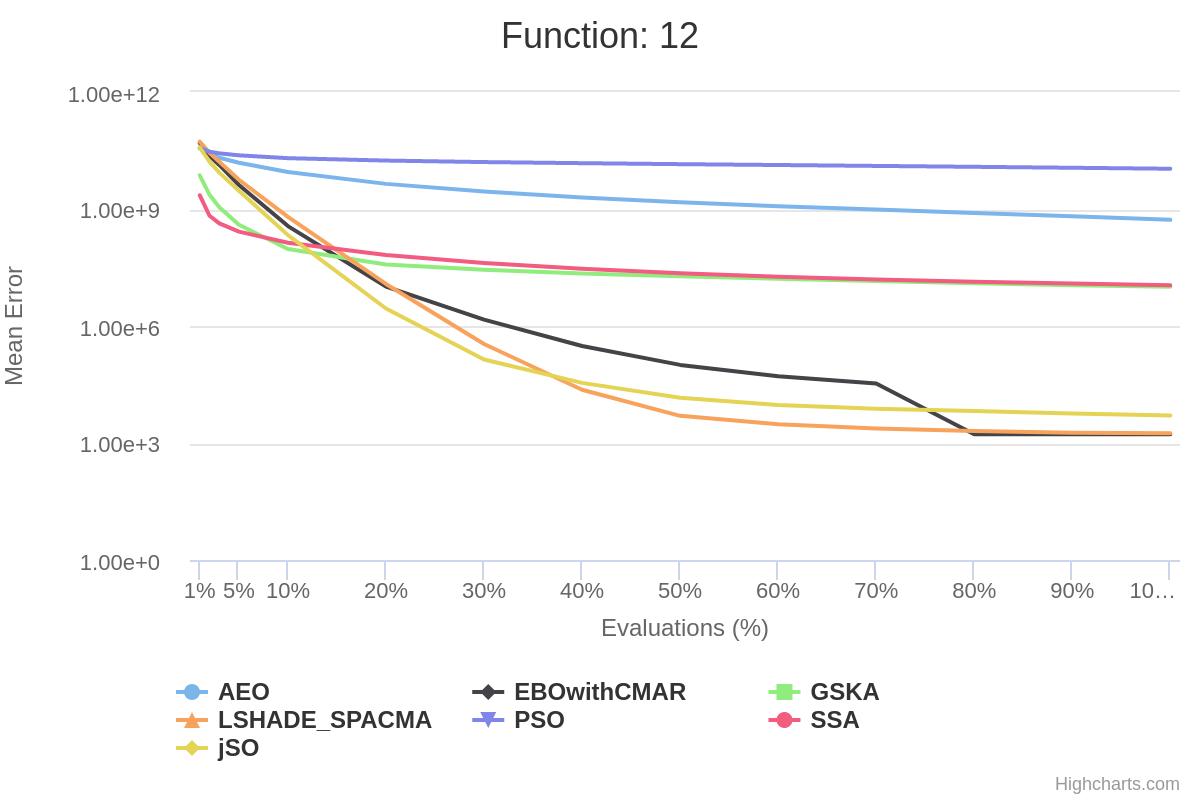}}\\
    \subfloat[\label{fig:conv_28}]{\includegraphics[width=0.7\linewidth]{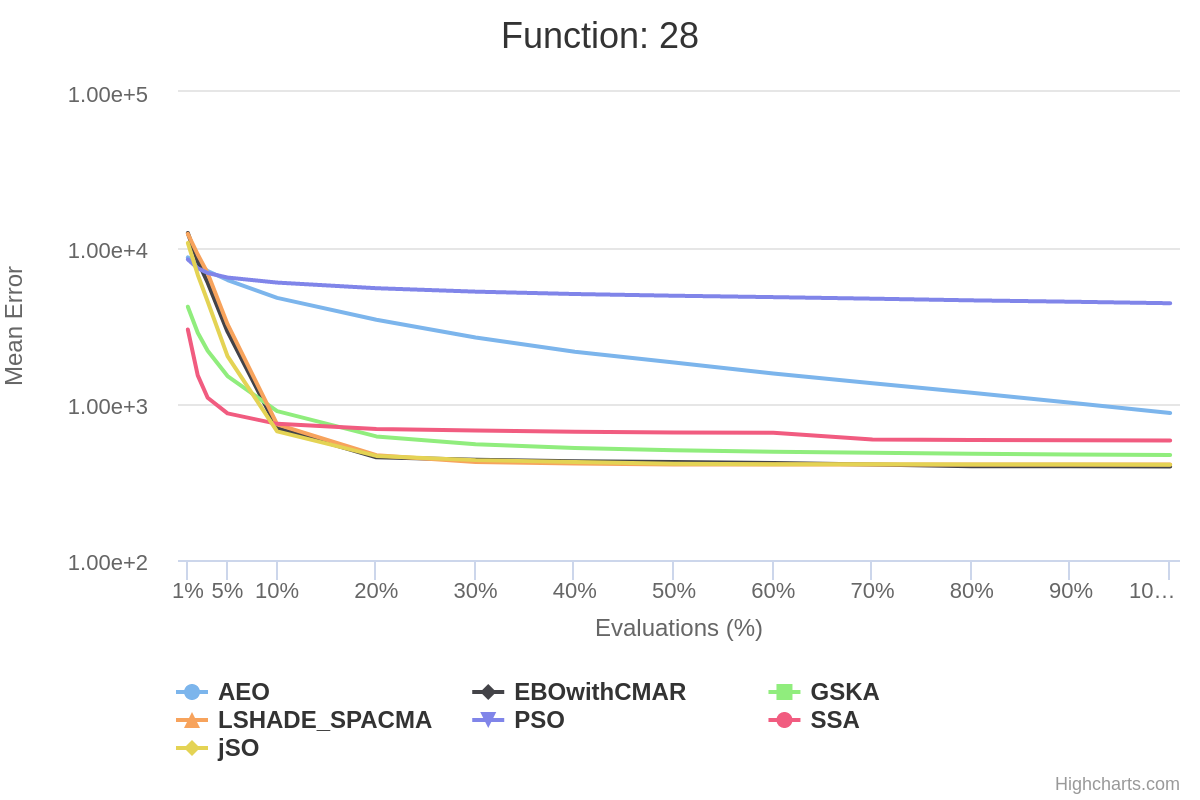}}
    \caption{Convergence curves for several functions of the CEC'2017 benchmark and dimension 10: (a) function 12; (b) function 28.}
    \label{fig:convs}
\end{figure}

Another typical visual representation is that of the convergence of an algorithm. In this case, the variable being discussed is the convergence speed of the methods involved in the comparison.
An example of this kind of plot is provided in Figure \ref{fig:convs}. As can be seen in Figure \ref{fig:conv_11}, which depicts the convergence speed of multiple algorithms on F11 of the CEC'2017 benchmark, although EBOwithCMAR is able to reach the minimum error among the competing algorithms, it is not until the end of the run that it overcomes other methods. For the majority of the time, jSO yields better results.

With a similar visual idiom, we can represent very different data. For instance, it could be interesting to measure the ratio of problems solved as the number of fitness evaluations increases. These plots not only evince which algorithm is able to reach the optima in more problems, but also how much effort is required to accomplish it. An example of this visualization is provided in Figure \ref{fig:coco}, which can be found in Section \ref{sec:case_study_real}.

\begin{figure}[!ht]
  \centering
  \subfloat[\label{fig:box_18}]{\includegraphics[width=0.5\linewidth]{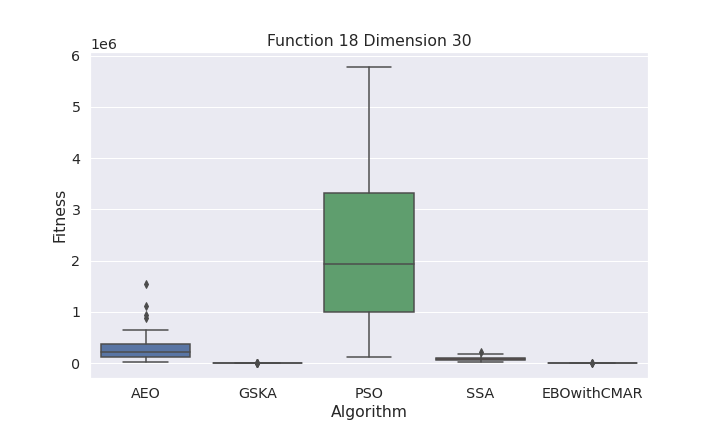}}
  \subfloat[\label{fig:box_26}]{\includegraphics[width=0.5\linewidth]{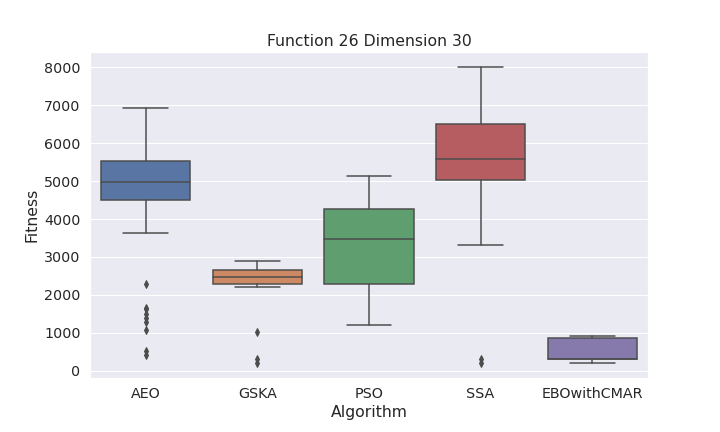}}
  \caption{Box-plots for the CEC'2017 benchmark and dimension 30: (a) function 18; (b) function 26.}
  \label{fig:boxes}
\end{figure}

Finally, even common visual representations such as boxplots can be very useful. In the context of bio-inspired optimization, it is a good way to show not only the mean error, but also the error for the different runs. Figure \ref{fig:boxes} provides two examples comparing the results of different algorithms for the case study in Section \ref{sec:case_study_real}.

The visual idioms suitable to represent information in this kind of comparisons are not limited to the two examples given in this section. There are many other alternative representations that can help to gain insight in the results under discussion.

What should be clear is that different idioms support different types of analyses, and that visualization techniques should be carefully selected in order to present the results in a summarized yet insightful fashion. All in all, our recommendation at this point is not only to visualize the results of the comparison, but also to use these techniques to complement and/or summarize the information provided by other means.

\section{Guideline \#3: Components Analysis and Parameter Tuning of the Proposal} \label{sec:guideline3}

This third guideline could be seen as a check-list for the discussion section. It covers the full proposal analysis, from the statement of the hypotheses to be proved by the experimentation to the presentation of the results of the different comparisons needed to assess the contribution of the work at hand. This section elaborates on this list by pausing at the following aspects: origin and work hypotheses that motivate the proposal (Section \ref{sec:guid4_origin}), the identification of the search phases, with claims on them solidly informed by empirical evidences (Section \ref{sec:guid4_explorvsexploit}), the individual analysis of algorithmic components of the proposal (Section \ref{sec:guid4_components}), and parameter tuning and analysis (Section \ref{sec:guid4_parameter}).

\subsection{Origin, hypotheses and proposal} \label{sec:guid4_origin}

Before starting to discuss about the benefits of the novel approach(es) proposed in the work, it is necessary to have a clear perspective of the expected results, i.e., the work hypotheses of the present study. Furthermore, authors should clearly describe how the proposal helps to attain the targeted objectives. Most of the times we assume that the main contribution of a work is a new algorithm or a new modification capable of improving the state of the art in some particular benchmark (properly chosen as already discussed in Section \ref{sec:guideline1_benchmark}). While this assumption often holds in practice, any other contribution to be taken into account must be clearly highlighted at this point for the work to be considered relevant. We will discuss further on this issue in Section \ref{sec:guideline4}.

\subsection{Search phases identification} \label{sec:guid4_explorvsexploit}

A key issue when solving optimization problems with complex fitness landscapes is to keep an appropriate balance between exploration and exploitation \cite{Herrera-Poyatos2017,LaTorre2010d}. This is a recurrent statement in many contributions, especially when the obtained results seem to support that hypothesis (at least in terms of overall accuracy). However, most of these studies fail to provide evidence on how the exploration/exploitation balance is maintained. It is normally not enough to state that \textit{``algorithm A is better than algorithm B because it properly maintains the exploration/exploitation balance''}. This type of statements requires an empirical analysis to check to which extent this claim is supported by evidence.

The work reported in \cite{Crepinsek2013} analyzes this issue from a dual perspective. First, they inspect how different authors measure the exploration/exploitation balance, to conclude that this is mainly carried out by means of indirect measures (e.g., the diversity of the solutions). Second, they propose a taxonomy of methods that aim to promote population diversity. Authors eager to include an analysis of this type in their research works should conduct a quantitative experimental study to justify the kind of statements that we have already mentioned. This can be achieved either by using evolutionary operators that explicitly enhance this balance \cite{epitropakis_balancing_2008, haq_novel_2019,hussain_trade-off_2020} or by updating the algorithm or the operators to take into account some kind of indirect measure \cite{vafaee_balancing_2014,LaTorre2014}.

\subsection{Components analysis and simplicity/complexity} \label{sec:guid4_components}

It is common in recent literature, especially in those papers where the proposal is evaluated on a well-known benchmark, that new proposals are built upon previous existing algorithms. Those new methods normally i) improve previous algorithms by updating or adding new characteristics to their baseline search procedure; or ii) combine existing methods to create a hybrid algorithm of some kind. However, few of these works analyze individually each of the improvements/components of the new proposal. This is an important issue from an algorithmic design perspective. It is true that \emph{powerful} algorithms are usually sought (in terms of their ability to find solutions as close as possible to the global optimum). But it is not less true that \emph{simplicity} should be considered as another preferential aspect in the design of new optimization techniques. Simplicity in algorithmic design has a number of advantages:
\begin{itemize}[leftmargin=*]
	\item Simple algorithms normally have less parameters to adjust (or, at least, less sensitive ones).
	\item Their behavior is more predictable, as there are less components involved.
  \item They can be described and implemented more easily. 
	\item It is less likely that the algorithm overfits one particular benchmark.
\end{itemize}

All these reasons are important enough to pay attention to the complexity of the new algorithm. For this reason, it is mandatory to provide an in-depth analysis of the contribution of each of the components of the new method to its overall performance. Every change or addition on top of the original algorithm must be supported by a significant contribution to the improved behavior of the novel method. Furthermore, if this contribution is shown to be small, this improvement should be considered for removal in the interest of simplicity. In this sense, we encourage the authors to use the same statistical validation methods described in Section \ref{sec:guideline2_stat} to compare each of the individual components of the algorithm. This is the same procedure that we recommend to compare the proposal with other state-of-the-art algorithms (see Section \ref{sec:case_study}). 

An illustrative example in this direction is \cite{Piotrowski2018}. This work analyzes one of the best performing algorithms of the IEEE CEC'2016 competition on real-parameter single objective optimization, namely, L-SHADE-EpSin \cite{Awad2016}. One of the conclusions of this analysis is that only one of the multiple additions to the base L-SHADE algorithm provides some significant improvement in the results (the initialization of the F parameter to $0.5$ during the first half of the search). The other modifications (materialized through the inclusion of several local search strategies) were found no significantly better. Moreover, they favored a bias in the search towards solutions around the origin of the search space, as also buttressed by \cite{Piotrowski2018}. This means that, even for competitive algorithms, the contribution of each component should be carefully evaluated. This is so, since a simplified version of the algorithm will always be easier to maintain, and can even lead to better results.

A second example aligned with our recommendations at this point emerges from the results of the MOS-SOCO2011 algorithm presented in \cite{LaTorre2010d}. By virtue of the Multiple Offspring Sampling (MOS) framework, this optimization technique combines two well-known algorithms: DE and the first local search method (MTS-LS1) of the Multiple Trajectory Search (MTS) algorithm \citep{Tseng2008}. The MOS-SOCO2011 hybrid algorithm was evaluated on the benchmark proposed for the Soft Computing Special Issue on the scalability of evolutionary algorithms and other metaheuristics for large-scale continuous optimization problems \citep{Lozano2011}. MOS-SOCO2011 obtained the best overall results among all the participants in the special issue. In \citep{LaTorre2010d}, authors reported not only the results for the proposed MOS-SOCO2011 algorithm, but also those of each of the independent components, DE and MTS-LS1, which are shown in Table \ref{table:1000d_res}. This threefold compilation of results allows for a direct comparison on the number of functions solved to the maximum precision (14, 1 and 4 for MOS-SOCO2011, DE and MTS-LS1, respectively), and sheds light on the synergy of both algorithms: except for functions Schwefel 2.21 and Schwefel 1.2, for which the MTS-LS1 algorithm obtained the best results, the hybrid method was able to reach the performance of the best one of its composing algorithms, normally outperforming them.

\begin{table}[ht!]
\caption{Results for MOS-SOCO2011, DE and MTS-LS1 over different 1000-D functions \cite{LaTorre2010d}}
\centering
\begin{tabular}{lccc}
\toprule
Benchmark function  & MOS-SOCO2011 & DE       & MTS-LS1  \\ \midrule
Sphere            & 0.00e+00     & 3.71e+01 & 1.15e-11 \\
Schwefel 2.21     & 4.25e-01     & 1.63e+02 & 2.25e-02 \\
Rosenbrock        & 6.15e+01     & 1.59e+05 & 2.10e+02 \\
Rastrigin         & 0.00e+00     & 3.47e+01 & 1.15e-11 \\
Griewank          & 0.00e+00     & 7.36e-01 & 3.55e-03 \\
Ackley            & 0.00e+00     & 8.70e-01 & 1.24e-11 \\
Schwefel 2.22     & 0.00e+00     & 0.00e+00 & 0.00e+00 \\
Schwefel 1.2      & 1.94e+05     & 3.15e+05 & 1.23e+05 \\
Extended f10      & 0.00e+00     & 6.26e-02 & 1.99e+03 \\
Bohachevsky       & 0.00e+00     & 1.67e-01 & 0.00e+00 \\
Schaffer          & 0.00e+00     & 4.42e-02 & 1.99e+03 \\
f12               & 0.00e+00     & 2.58e+01 & 5.02e+02 \\
f13               & 8.80e+01     & 8.24e+04 & 8.87e+02 \\
f14               & 0.00e+00     & 2.39e+01 & 2.23e+03 \\
f15               & 0.00e+00     & 2.11e-01 & 0.00e+00 \\
f16               & 0.00e+00     & 1.83e+01 & 1.00e+03 \\
f17               & 2.25e+01     & 1.76e+05 & 1.56e+03 \\
f18               & 0.00e+00     & 7.55e+00 & 1.21e+03 \\
f19               & 0.00e+00     & 2.51e-01 & 0.00e+00 \\ \midrule
Solved functions & 14           & 1        & 4        \\
\bottomrule
\end{tabular}
\label{table:1000d_res}
\end{table}

Additionally, a statistical comparison was also carried out, reporting significant p-values for both comparisons (MOS-SOCO2011 versus DE, and MOS-SOCO2011 versus MTS-LS1). These two comparisons altogether provide enough confidence to support the superiority of the hybrid method over each of its composing algorithms.
This is not the only example of such a comparison. Authors of the L-SHADE algorithm follow a similar approach in \cite{Tanabe2014}, comparing the new version of their algorithm to previous ones, and evaluating the addition of new components to prove their benefits.

\subsection{Parameter tuning and analysis} \label{sec:guid4_parameter}

One important problem in the design of an algorithm is the number of free parameters that can be adjusted to modify its behavior. In general, the more flexibility, the more control parameters to adjust. Very often, the values selected for these parameters are so determinant in the search that even a well designed algorithm can yield bad results with the wrong parameter values. As a consequence, the selection of the values for these internal parameters is a critical decision that should not be underestimated nor overseen.

Unfortunately, the selection of the right parameters values or the adaptation mechanism is not an easy task, because most of the times there is not a clear criterion to guide that selection, and it could be considered an optimization problem itself.
Parameter tuning can be tackled in different ways:

\begin{itemize}[leftmargin=*]
  \item \textbf{Offline vs. online tuning}. \textit{Offline} tuning approaches search the optimal EA parameter settings for several representative problems. Then, these values are applied to new problems \cite{dejongParameterSettingEAs2007}, which means that the tuning is carried out before the algorithm is run on the problem to be solved. The main drawback of this approach is that the values obtained do not necessarily need to be optimal for new problems, but, in practice, robust parameters configurations can be obtained. An alternative approach would be to conduct the parameter tuning during the run. This strategy, in theory, can obtain better adapted values to each problem, but it is a more complex method to implement.

  \item \textbf{Static, adaptive and self-adaptive parameter control.} In \cite{eibenNewWaysCalibrate2008}, parameter control approaches (those carried out during the run of the algorithm) are divided into one of the three following categories: deterministic, adaptive and self-adaptive. Deterministic parameter control means that the value of the strategy parameter is updated following some deterministic rules, without feedback from the search. Adaptive parameter control takes place when some feedback from the search process is used to determine the direction and/or the magnitude of the adjustment of the parameters. Finally, self-adaptive parameter control happens when the values of the parameters are embedded within the solution to optimize. Thus the parameter setting is implicit during the search. Several successful examples of self-adaptation are discussed below.
\end{itemize}

The most usual approach to deal with the tuning problem is to do it offline and experimentally, comparing the results obtained on multiple combinations of the parameters values. However, there are several pitfalls in which a researcher may fall when conducting a manual tuning:
\begin{itemize}[leftmargin=*]
\item \emph{Test each parameter on a small number of predefined values, without taking any feedback into consideration}: some works only test extreme values of the parameter(s) to be tuned (either minimum or maximum over its range). Under these circumstances, the number of values to test should be extended to check whether a better result can be obtained with other values over the range of the parameter(s).

\item \emph{Tune only a small part of their parameters}: in this case, the values of the remaining parameters are guessed or initialized to fixed values, without analyzing which parameters influence most on the performance of the algorithm. This analysis would eventually justify which parameters should be selected and carefully tuned, but it is rarely done in the literature.

\item \emph{Try to tune each parameter independently, keeping the others fixed}: this widely adopted approach poses many problems. The results obtained by varying just one parameter depend largely on the values given to the others, since it is not uncommon that multiple parameters influence on each other. Therefore, all value combinations of parameters should be optimized altogether. While it is true that exploring all the possible combinations can lead to a combinatorial explosion, there are several techniques in the field of experimental design \citep{stanley66_influen_fisher_desig_exper_educat} that can alleviate this problem, such as the fractional design \citep{lundstedt98_exper_desig_optim} or the alternative Taguchi methods \citep{taguchi}.

\item \emph{Use parameter values tuned by other authors in previous experiments}: these parameters were usually tuned for a different problem, and thus their values might not be appropriate for the problem/benchmark under consideration. Of course, these values can still be used, but they should never be considered to be the optimal ones, as their competitiveness on a different benchmark can not be guaranteed \cite{wessingParameterTuningBoosts2010a,MolinaTuned}.

\item \emph{Lack of statistical tests in the comparisons to tune the parameters values of the proposal}: due to the stochastic nature of meta-heuristic algorithms, selecting parameter values based on average fitness values is not enough. The same statistic procedures used to compare multiple techniques should be used when comparing different configurations of an algorithm. A choice of the most promising values is at discretion of prospective authors, but the use of statistical tests should be always enforced, no matter which parameter tuning approach is being carried out. In fact, as shown in \citep{Garcia2014}, the use of statistical tests for this kind of comparisons can be straightforward.
\end{itemize}

An alternative approach to parameter tuning is the self-adaptation of the parameters \citep{eiben99_param_contr_evolut_algor}. Under this approach, parameters are not given a fixed value. Instead, a process is devised to automatically adjust their value according to the feedback obtained from the optimization process. This type of adaptation mechanism has been successfully applied to DE, which is very sensitive to its parameters \cite{brest06_self_adapt_contr_param_differ_evolut,neri09_recen_advan_differ_evolut}. The work in \cite{teo17_scalab_analy_fixed_versus_self} showed the convenience of self-adaptive values versus fixed parameter settings. In \cite{qin09_differ_evolut_algor_with_strat} the first DE with self-adaptive parameters was presented, drawing parameter values by sampling a distribution which mean is updated by considering new solutions entering the population. Since then, new algorithms improving the self-adaptation mechanism \cite{JADE,SHADE2013} have been proposed. In \cite{Tanabe2014}, the previous self-adaptive parameters were complemented with an adaptive population size that linearly decreases along the search process. In \cite{brest17_jso} a tuning of parameters was applied, improving further the results of the overall algorithm.

Although adaptive parameters are a clear improvement over fixed parameters, both in terms of ease of usage and robustness, not all the parameters can be self-adjusted in this way (in particular, some more internal parameters). Thus, even in self-adaptive algorithms, there are fixed parameters that must be tuned to improve the results even more (i.e., \cite{brest17_jso}).

Fortunately, the tuning of parameters can be automatically carried out. There are several useful and consolidated tools of this type with different features \citep{tuning_method}. In the following paragraphs we briefly describe some of the state-of-the-art methods:
\begin{itemize}[leftmargin=*]
\item Sequential Parameter Optimization, SPO~\cite{bartz-beielsteinSequentialParameterOptimization2005}, is a heuristic that uses Latin Hypercube Sampling to determine the parameter values keeping the computation cost low.

\item Iterated Racing for Automatic Algorithm Configuration, IRACE \cite{irace}, is an optimization algorithm that uses different runs (races) and statistical testing to identify combinations of parameters that are worse than others, providing an automatic process to optimize a variety of parameters.

\item Relevance Estimation and Value Calibration of Evolutionary Algorithm Parameters, REVAC \citep{REVAC}, is an evolutionary algorithm that uses multi-parent crossover and entropy measures to estimate the relevance of parameters.
  
\item ParamILS \cite{ParamILS}, is a versatile stochastic local search approach for automated algorithm configuration.
\end{itemize}

Although I-RACE has yielded better results for us in the past, all the aforementioned alternatives are robust and consolidated tools, so the selection of one tool over the others will depend on the problem features and personal preferences.

To conclude with this issue, we would like to make an additional remark. The tuning of the parameters of a new proposal can have an important impact in the objectivity of the comparison. This situation can occur when the new proposal is the only algorithm which parameters are tuned for the particular benchmark used in the experimentation, whereas reference algorithms use parameters values proposed by their respective authors under different experimental conditions. This will probably mean that those algorithms are not expected to report very good results for the new experimental scenario. In this case, having the parameters of the new proposal tuned for the benchmark considered in the work could give our proposal an unfair advantage over the other ones, generating a bias in favor of the proposed algorithm in the comparison. Ideally, the solution should be to compare tuned versions of all the algorithms \citep{MolinaTuned}, but the cost of doing that could become too expensive and computationally unaffordable. However, when the algorithms are compared against standard benchmarks (not defined \emph{ad-hoc} for each paper), this risk is minimized, because all the algorithms were tested under the same experimental conditions.

\section{Guideline \#4: Why is my algorithm useful?}
\label{sec:guideline4}

The final step of a successful proposal is a thorough discussion of the results. This discussion must answer a crucial question: why is my algorithm useful?

The most obvious answer to this question is \emph{``because it outperforms current state-of-the-art methods''}. If the algorithm falls within this first category of proposals, and if this outperforming behavior is validated by principled means (as those shown in this manuscript), the contribution has a clear scientific value and can be contributed to the community in the form of a publication. However, this is not always the case, but it does not mean that the contribution is not significant.

There are a number of reasons to accept a new proposal even if it is unable to outperform the best-so-far algorithms. Nonetheless, under these circumstances it is even more important the discussion of the results. The benefits of adopting the method proposed in such a contribution should be clearly stated and highlighted accordingly. We next discuss on some of the reasons that can be considered enough for a new proposal to be accepted:

\begin{itemize}[leftmargin=*]
	\item The first of these reasons is the quality of the results. If, as mentioned before, the results clearly outperform current state-of-the-art methods, the authors have a solid argument for their paper to be accepted. Sometimes, it could be enough that the results are particularly good for a subset of the problems, given that this behavior can be identified and characterized. This does not imply that the rest of the guidelines provided in our paper can be neglected. The discussion of the results should be rigorous and the conclusions should be clearly presented, without any ambiguity nor vagueness.

	\item The second of these reasons is novelty: if a newly proposed algorithm has the potential to evolve and become competitive with current state-of-the-art methods, it should be presented to the community. Nonetheless, special attention should be paid at this point to avoid the problems described by \cite{Sorensen2015,del2019bio}: it is absolutely mandatory that, besides the bio-inspired metaphor, the new algorithmic proposal is competitive enough for a set of problems. Furthermore, we firmly advocate for the development of a unified description language for meta-heuristic algorithms, capable of unambiguously describing each of the algorithmic steps of new proposals, leaving aside any metaphorical language. We utterly believe that efforts in this direction should be intensified, building upon the initial postulations established in some recent works \cite{swan2015research,swan2020towards}. Specifically, meta-heuristics components (including search operators and algorithmic behavioral patterns) and interfaces between them should be standardized towards objectively assessing similarities and differences between metaphor-based solvers \cite{lones2020mitigating}. A novel metaphor is by no means a sufficient guarantee for a significant scientific contribution. 

	\item The third of these reasons is methodological, i.e., the relevance of some of the building blocks of the overall algorithm. A particular algorithm can include a given component (for example, a local optimizer) that can be of relevance even if the algorithm as a whole is not completely competitive with respect to the prevailing literature. A good example supporting this claim can be observed in co-evolutionary frameworks, which usually include a procedure to identify the subcomponents that will be individually co-evolved. In those cases, even if the subcomponent optimizer is not very sophisticated, the co-evolutionary framework can be relevant by itself. In this sense, it is important to select the appropriate framework to highlight the desired characteristic of the proposed algorithm, as discussed in Section \ref{sec:guideline1_benchmark}. Following the same example of subcomponents identification, a researcher focused on large-scale global optimization should consider the CEC'2013 benchmark that explicitly studies this issue \cite{Li2013a}. Nevertheless, this is a quite subjective consideration, so authors should clearly highlight these benefits to avoid debatable claims.
\end{itemize}

\section{Case Studies}\label{sec:case_study}

In order to exemplify the application of the previous guidelines, this section elaborates on several case studies that follow our proposed methodology. In particular, we consider the following two scenarios:
\begin{itemize}[leftmargin=*]
    \item The study of different existing algorithms, wherein the goal is to analyze the advantages and drawbacks of each one of them. In this first scenario, the priority is to compare all the methods following different criteria, towards determining under which circumstances each of them can be recommended.

    \item The proposal of a new algorithm, in which the priority is to provide informed evidence of the competitiveness of the new proposal, as well as its advantages over previous algorithms.
\end{itemize}

The structure of this section conforms to the above scenarios. First, in Subsection \ref{sec:case_study_real} we present several recent bio-inspired algorithms for real-parameter optimization, and subsequently we resort to our proposed guidelines so as to compare them fairly. Then, in Subsection \ref{sec:case_study_lsgo} we discuss on the second case of study, where a new algorithm is proposed for a specific type of optimization problem: large-scale global optimization.

\subsection{Several modern bio-inspired algorithms for real-parameter optimization}\label{sec:case_study_real}

In this first case study we simulate the scenario in which a comparison of several bio-inspired algorithms is designed for real-parameter optimization. There are two possible reasons for which this comparison can be tackled. On the one hand, we might be willing to propose a new solver, for which we must assess the performance of existing algorithms to use them as a reference for our proposal. On the other hand, we might be interested in solving a particular problem similar to the ones considered in the comparison, for which we analyze several algorithmic options in order to ascertain which one to use.

The algorithms considered in this first case study are recent methods presented in top-tier journals:
\begin{description}
    \item[Squirrel Search algorithm (SSA):] This is a bio-inspired algorithm that imitates the foraging behavior of squirrels. It divides the solutions into three groups based on their fitness (the best one, the three next best ones, and the remaining ones), and adopts different criteria to combine them considering this grouping arrangement, combining each solution with a solution in a superior category by a random lineal combination \cite{jainNovelNatureinspiredAlgorithm2019a}.
    
    \item[Gaining-sharing knowledge based algorithm (GSKA):] This is an algorithm inspired by the human behavior when sharing knowledge. It has been observed that there is more acceptation in the early phases, and that people get more questioning in later stages. For an algorithmic point of 
    view, it uses two different criterion to optimize each dimension: in the first one, referred to as \emph{junior gaining-sharing}, the variable at hand is updated considering the variables of its immediate better and worse individuals in the population. In the second criterion, \emph{senior gaining-sharing}, the 
    variable is updated considering the best and worst individuals in the population. Initially all variables are updated by the junior gaining-sharing mechanism, and during the search, an increasing number of variables are updated by the senior gaining-sharing criterion \cite{mohamedGainingsharingKnowledgeBased2020}.
    
  \item[Artificial Ecosystem-Based Optimization (AEO):] This is a nature-ins\-pired meta-heuristic that finds its motivation in the flow of energy through an ecosystem. In this proposal, the population is updated by means of an iterative process composed by different phases. The first phase is \textit{production}, in which one solution undergoes a small and decreasing random update. The second one is \textit{consumption}, in which every individual is randomly classified as \emph{herbivore}, \emph{omnivore}, or \emph{carnivore}. Depending on its category, the individual is updated as per a different rule. Newly produced solutions are inserted into the population if they improve the previous ones (replacing them). In the final phase (\textit{decomposition}), each individual is mutated by a dispersion equation \cite{zhaoArtificialEcosystembasedOptimization2020}.

  \item[Enhanced LSHADE-SPACMA Algorithm (ELSHADE-SPACMA):] This is a new hybrid algorithm that alternates, at each iteration, i) the application of a DE strategy that considers, in its mutation operator, not only the best individuals but also the worst ones \cite{mohamedAdaptiveGuidedDifferential2019}, with ii) an improved LSHADE-SPACMA \cite{mohamedLSHADESemiparameterAdaptation2017} that adapts its \textit{p} parameter to enforce exploitation in its final stages \cite{hadiSingleObjectiveRealParameterOptimization2021}. LSHADE-SPACMA is a hybrid algorithm that applies, with a certain probability, one of the well-known algorithms L-SHADE \cite{Tanabe2014} and CMA-ES \cite{Hansen2001}. The probability of applying each algorithm is adapted by considering the improvements obtained by each of them over the search. In its seminal paper, ELSHADE-SPACMA is stated to outperform previous winners of real-parameter optimization competitions.
\end{description}

    Following our suggestions, we have used existing implementations of the algorithms, in particular, for AEO and GSKA we have used the Mealpy software\footnote{\url{https://github.com/thieu1995/mealpy}}. For SSA we have used the implementation of the Indago project\footnote{\url{https://pypi.org/project/Indago/}}. Finally, for ELSHADE-SPACMA we have used the source code provided by the authors of the algorithm\footnote{\url{https://sites.google.com/view/optimization-project/files}}.

In the comparison study discussed throughout this section, we follow the guidelines proposed in this work so as to properly conduct the experiments with these algorithms and other reference methods. During this study, we also discuss the advantages of our methodological proposal. We next start by the first guideline.

\subsubsection{Selecting the benchmark as per Guideline \#1}

First, we must choose an adequate benchmark for measuring the performance of the considered algorithms. Following the recommendations of Guideline \#1 described in Section \ref{sec:guideline1}:
\begin{itemize}[leftmargin=*]
    \item We must properly select the benchmark: without any unexpected bias, with the right level of complexity, and for the type of problem addressed by the algorithm.

    \item We must enforce the usage of a standard benchmark that fulfills the previous requirements, considering the characteristics of the problems for which the algorithms were originally designed.
\end{itemize}

In this case, all these algorithms have been especially designed for real-parameter optimization. Fortunately, there are several benchmarks well-designed for this type of problems, known to avoid unexpected bias, and endowed with different levels of complexity. In particular, several benchmarks have been proposed within the context of real-parameter optimization, both in the IEEE CEC and GECCO competitions. For our case study, we embrace two of these benchmarks: the CEC'2017 \cite{funcec2017real} and the COCO benchmarks \cite{nikolaus_hansen_2019_2594848}, for the following reasons:
\begin{itemize}[leftmargin=*]
    \item There are several benchmarks, and all of them are equally good choices. However, some of the benchmarks are rather complementary as they follow different approaches. The real-parameter CEC benchmarks are more focused on proposing difficult objective functions and measuring the final error obtained by each algorithm. On the other hand, the COCO benchmark is made up of simpler functions to allow measuring the performance of an algorithm from different perspectives: how many problems each algorithm is able to solve, how many objective function evaluations every compared solver requires for solving each problem, and other aspects alike.
    
    \item When the objective is to compare several algorithms and to analyze them fairly, it is suitable to have several sources of information that allow us to confirm our insights and compare the algorithms from different points of view.
\end{itemize}

\subsubsection{Selecting the performance measure as per Guideline \#1}

Another important decision to make is the choice of an adequate performance measure. As discussed in Guideline \#1, although the final value of the fitness error is a popular choice, it is not the only alternative to compare the performance of several algorithms. Another interesting option is, for example, the analysis of the evolution of the fitness value as the algorithms in the benchmark iterate to solve each problem.

In our case, the two selected benchmarks are focused on different yet complementary measures:
\begin{itemize}[leftmargin=*]
    \item In the CEC'2017 benchmark, the error in reported for different milestones , with different ratios of evaluations: 1\%, 2\%, 3\%, 5\%, 10\%, 20\%, 30\%, 40\%, 50\%, 60\%, 70\%, 80\%, 90\%, and 100\%. In our study we aim at showing the evolution of the algorithms as the number of function evaluations increases. Consequently, we select several of these milestones when reporting the results attained by each algorithm in the comparison.
    
    \item Additionally, we pay special attention to the 100\% milestone towards providing a final comparison score of the results, without downplaying the importance of reaching competitive fitness values with fewer evaluations.
    
    \item In the COCO benchmark, the performance is measured in a different fashion. In particular, the ratio of functions for which the optimum has been reached is computed as the number of objective function evaluations increases.
\end{itemize}

We consider that with these three measures we will obtain a relatively complete view of the performance of the different algorithms included in our study.

Finally, the results are gathered for every function, independently. However, for a more general analysis, we are also going to use aggregation, considering the \emph{average ranking}, which is calculated by sorting the algorithms for each function based on their error (assigning lower rankings to better algorithms). Then, the average ranking is computed so that an algorithm with a lower average ranking value is declared to perform better, on average, than another one with a higher ranking value.

\subsubsection{Selecting the reference algorithms as per Guideline \#1}
\label{sec:case1_guideline1_reference}

In order to conduct a fair comparison, a clear criterion is needed to select other algorithms to be included in the benchmark, grounded on the need for assessing the convenience of the considered solvers with regards to its competitive performance when compared with prevailing methods. Following our guidelines, we should:
\begin{itemize}[leftmargin=*]
    \item \textit{Compare against reference algorithms}: The idea is to select a well-known method to compare whether the algorithms perform competitively against it. For this purpose we have considered a classic algorithm: Particle Swarm Optimization (PSO).
    
    \item \textit{Compare against similar algorithms}: this decision is especially relevant when an new algorithm is proposed based on particular methods or with well-defined characteristics that resemble those of other existing algorithms. However, in this use case the algorithms to be evaluated have very different sources of inspiration and different algorithmic behavioral patterns. Thus, we consider that there is no need for selecting other similar algorithms in addition to PSO.
    
    \item \textit{Compare against competitive algorithms}: this is a hard decision to make, since it is often difficult to scrutinize the entire state-of-the-art related to the optimization problem/algorithm/benchmark under consideration. However, since the considered benchmarks were used in several competitions, we can easily find out competitive algorithms for each of them. In particular, the COCO benchmark tools \footnote{Available at: \url{https://github.com/numbbo/coco}} always compare against a previous winner of the competition. In the case of the CEC'2017 benchmark, we compare against the two best algorithms of the competition: jSO \cite{jSO} and EBOwithCMAR \cite{EBOwithCMAR}.
\end{itemize}

To summarize, in this case study we compare each algorithm: (1) to each other; (2) to a well-know reference algorithm (PSO); and (3) to three competitive algorithms (jSO and EBOwithCMAR for the CEC'2017 benchmark, and the winner of 2009 for the COCO benchmark).

\subsubsection{Experimenting and validating the results as per Guideline \#2}\label{sec:cec17_results}

Once the design of the experimentation is set up, the actual experiments are carried out, and results are validated. Following the recommendations on the use of statistical validation provided in Guideline \#2  (see Section \ref{sec:guideline2_stat}), normality and homocedasticity should be checked before the appropriate statistical test can be selected. However, in \cite{garcia08_study_use_non_param_tests} the results on a previous similar benchmark were analyzed for this two hypotheses. Such results clearly indicated than none of the assumptions were satisfied. For this reason, we have decided to use non-parametric tests. Furthermore, as also suggested by Guideline \#2, we have also applied Bayesian tests to compare the best performing approaches to each other.

\paragraph{CEC'2017 benchmark}

The first step suggested in Guideline \#2 is to calculate the average ranking of the algorithms, followed by non-parametric hypothesis testing. In order to realize this comparison, we resort to Tacolab \cite{tacolab}\footnote{Tacolab website: \url{https://tacolab.org/}}, a web tool that eases the comparison of algorithms with different criteria.
\begin{table}[ht]
  \centering
  \caption{Average ranking of the algorithms for each dimension}\label{table:ranking_cec}
  \vspace{2mm}
  \begin{tabular}{lccccc}
    \toprule
    Algorithm   & D10            & D30            & D50   & D100 & Mean \\
    \midrule
    ELSHADE-SPACMA  & 2.667          & 2.133          & \textbf{1.783}       & \textbf{1.383}       & \textbf{1.9915}\\
    EBOwithCMAR     & \textbf{1.933}       & \textbf{1.917}       & 2.167          & 2.367          & 2.0960\\
    jSO             & 2.317          & 2.050          & 2.117          & 2.583          & 2.2667\\
    GSKA            & 3.883          & 4.200          & 4.433          & 4.533          & 4.2623\\
    PSO             & 5.367          & 5.867          & 6.000          & 6.400          & 5.9085\\
    SSA             & 5.733          & 5.667          & 5.333          & 4.733          & 5.3665\\
    AEO             & 6.100          & 6.167          & 6.167          & 6.000          & 6.1085 \\
    \bottomrule
  \end{tabular}
\end{table}

Table \ref{table:ranking_cec} shows the average ranking for the CEC'2017 benchmark. Several observations can be made over the results in this table:
\begin{itemize}[leftmargin=*]
    \item Only ELSHADE-SPACMA performs best than previous winners of this competition: jSO and EBOwithCMAR. This is particularly the case for higher dimension problems (D50 and D100). None of the other recently proposed algorithms are competitive against these older reference algorithms. This is an interesting result, since the papers in which these new algorithms were first proposed do not compare against state-of-the-art methods. Instead, classic algorithms are just considered. This is even the case of GSKA, which was tested over the CEC'2017 benchmark, but did not include jSO nor EBOwithCMAR in the experiments.
    \item Among the recent proposals, ELSHADE-SPACMA clearly performs best, even better than previous winners: jSO and EBOwithCMAR.
      
    \item Apart from the aforementioned ones (ELSHADE-SPACMA, EBOwithCMAR and jSO), the proposal with the best average performance in this benchmark is GSKA, followed by SSA. We highlight that these algorithms do not result from the hybridization of previous ones.
    
    \item The reference algorithm PSO obtains worse results than most of the algorithms, except for AEO.
\end{itemize}

A full picture of these results can be displayed if we also measure how the performance of the different algorithms evolves during the search. Fortunately, the experimental conditions of the benchmark require that the error must be measured at different milestones: 1\%, 2\%, 3\%, 5\%, 10\%, 20\%, 30\%, 40\%, 50\%, 60\%, 70\%, 80\%, 90\%, and 100\% of the fitness evaluations.
\robustify\bfseries
\sisetup{round-mode=places, round-precision=2,round-half=even, detect-all}
\begin{table}
  \centering
  \caption{Evolution of the average ranking with regards to fitness evaluations in the CEC'2017 benchmark (dimension 10).}
  \label{tab:evol_cec_d10}
  \input{results_evol_cec_d10.tex}
\end{table}

\begin{table}
  \centering
  \caption{Evolution of the average ranking with regards to fitness evaluations in the CEC'2017 benchmark (dimension 30).}
  \label{tab:evol_cec_d30}
  \input{results_evol_cec_d30.tex}
\end{table}

\begin{table}
  \centering
  \caption{Evolution of the average ranking with regards to fitness evaluations in the CEC'2017 benchmark (dimension 50).}
  \label{tab:evol_cec_d50}
  \input{results_evol_cec_d50.tex}
\end{table}

\begin{table}
\centering
\caption{Evolution of the average ranking with regards to fitness evaluations in the CEC'2017 benchmark (dimension 100).}
\label{tab:evol_cec_d100}
\input{results_evol_cec_d100.tex}
\end{table}

Tables \ref{tab:evol_cec_d10}, \ref{tab:evol_cec_d30}, \ref{tab:evol_cec_d50} and \ref{tab:evol_cec_d100} show the average ranking of each method at these different milestones for dimensions 10, 30, 50, and 100, respectively. From these tables the following conclusions can be drawn:
\begin{itemize}[leftmargin=*]
 
    \item For dimensions 30 and 50, GSKA is a much faster algorithm and it is clearly better than all the other proposals up to 10\% of the fitness evaluations, including the previous winners of CEC competitions (EBOwithCMAR and jSO) and the most competitive recent algorithm (ELSHADE-SPACMA).

    \item EBOwithCMAR is the algorithm with the best results for dimensions 10 and 30 (very close to ELSHADE-SPACMA in those dimensions), whereas ELSHADE-SPACMA is the best performing algorithm for dimensions 50 and 100, closely followed by EBOwithCMAR and jSO.

    \item For dimensions 30, 50 and 100, ELSHADE-SPACMA is the best one since 40\% of the budget of evaluations. Although Table~\ref{table:ranking_cec} indicates that EBOwithCMAR obtains the best final results, ELSHADE-SPACMA achieves better results during most of the search, being only improved by the former at the end.

    \item Deciding which algorithm should be applied to a specific problem strongly depends on the effort that can be devoted to the search. In this benchmark, GSKA is better when less evaluations are allowed, whereas ELSHADE-SPACMA is preferred when a higher number of evaluations can be afforded.

\end{itemize}

Additionally, following Guideline \#2 (Section \ref{sec:guideline2}) we have conducted a statistical validation of the results to reject the hypothesis that the differences observed in the performance of the algorithms is due to their stochastic nature and not to actual differences in their performance. First, we use the Friedman rank-sum test to find out if significant differences can be found among all the algorithms. The p-values reported by this test are 3.61e-08, 1.51e-07, 7.58e-7, and 9.62e-10 for dimensions 10, 30, 50, and 100, respectively. Since all the p-values values are clearly lower than the $\alpha=0.05$ confidence level, we can state that differences among the algorithms exist and that are significant.
\begin{table}[ht]
  \centering
  \caption{Statistical validation for the CEC'2017 benchmark and dimension 10 (EBOwithCMAR is the control algorithm).}
  \vspace{2mm}
  \label{table:validation_cec17_d10}
  \begin{tabular}{lcl}
  \toprule
  EBOwithCMAR versus & Wilcoxon p-value & Wilcoxon p-value$*$\\
    \midrule
    AEO            & 2.702e-06 & 1.621e-05 ${\surd}$ \\
    SSA            & 3.703e-06 & 1.852e-05 ${\surd}$ \\
    PSO            & 4.110e-06 & 1.852e-05 ${\surd}$ \\
    GSKA           & 4.374e-05 & 1.312e-04 ${\surd}$ \\
    ELSHADE-SPACMA & 0.027     & 0.055           \\
    jSO            & 0.225     & 0.225           \\ 
    \bottomrule
    \noalign{\vskip 2mm}   \multicolumn{3}{l}{$\surd$: statistical differences exist with significance level $\alpha=0.05$.}\\
    \multicolumn{3}{l}{$^*$: p-value corrected with the Holm procedure.}\\&&
  \end{tabular}
\end{table}

\begin{table}[ht]
  \centering
  \caption{Statistical validation for the CEC'2017 benchmark and dimension 30 (EBOwithCMAR is the control algorithm).}
  \vspace{2mm}
  \label{table:validation_cec17_d30}
  \begin{tabular}{lcl}
    \toprule
    EBOwithCMAR versus  & Wilcoxon p-value & Wilcoxon p-value$*$\\
    \midrule
    AEO            & 2.702e-06 & 1.621e-06 ${\surd}$ \\
    PSO            & 2.702e-06 & 1.621e-06 ${\surd}$ \\
    SSA            & 2.702e-06 & 1.621e-06 ${\surd}$ \\
    GSKA           & 4.110e-06 & 1.621e-06 ${\surd}$ \\
    ELSHADE-SPACMA & 0.226     & 0.452\\
    jSO            & 0.431     & 0.452\\
        \bottomrule
    \noalign{\vskip 2mm}   \multicolumn{3}{l}{$\surd$: statistical differences exist with significance level $\alpha=0.05$.}\\
    \multicolumn{3}{l}{$^*$: p-value corrected with the Holm procedure.}\\&&
  \end{tabular}
\end{table}

\begin{table}[ht]
  \centering
  \caption{Statistical validation for the CEC'2017 benchmark and dimension 50 (ELSHADE-SPACMA is the control algorithm).}
  \vspace{2mm}
  \label{table:validation_cec17_d50}
  \begin{tabular}{lcl}
    \toprule
    ELSHADE-SPACMA versus & Wilcoxon p-value & Wilcoxon p-value$*$\\
    \midrule
    AEO         & 1.863e-09 & 1.118e-09 ${\surd}$ \\
    GSKA        & 1.863e-09 & 1.118e-09 ${\surd}$ \\
    PSO         & 1.863e-09 & 1.118e-09 ${\surd}$ \\
    SSA         & 1.863e-09 & 1.118e-09 ${\surd}$ \\
    jSO         & 0.054     & 0.107 \\
    EBOwithCMAR & 0.509     & 0.509  \\
    \bottomrule
    \noalign{\vskip 2mm}   \multicolumn{3}{l}{$\surd$: statistical differences exist with significance level $\alpha=0.05$.}\\
    \multicolumn{3}{l}{$^*$: p-value corrected with the Holm procedure.}\\&&
  \end{tabular}
\end{table}

\begin{table}[ht]
  \centering
  \caption{Statistical validation for the CEC'2017 benchmark and dimension 100 (ELSHADE-SPACMA is the control algorithm).}
  \vspace{2mm}
  \label{table:validation_cec17_d100}
  \begin{tabular}{lcl}
    \toprule
    ELSHADE-SPACMA versus & Wilcoxon p-value & Wilcoxon p-value$*$\\
    \midrule
    AEO         &  3.725e-09 &  2.235e-08 ${\surd}$ \\
    GSKA        &  3.725e-09 &  2.235e-08 ${\surd}$ \\
    PSO         &  3.725e-09 &  2.235e-08 ${\surd}$ \\
    SSA         &  5.588e-09 &  2.235e-08 ${\surd}$ \\
    jSO         &  2.254e-05 &  4.507e-05 ${\surd}$ \\
    EBOwithCMAR & 3.128e-04  &  3.128e-04 ${\surd}$ \\
    \bottomrule
    \noalign{\vskip 2mm}   \multicolumn{3}{l}{$\surd$: statistical differences exist with significance level $\alpha=0.05$.}\\
    \multicolumn{3}{l}{$^*$: p-value corrected with the Holm procedure.}\\&&
  \end{tabular}
\end{table}

Once that significant differences are detected, we proceed with a multiple comparison, as the use of the Holm procedure keeps the family-wise error rate under control. Tables \ref{table:validation_cec17_d10}, \ref{table:validation_cec17_d30}, \ref{table:validation_cec17_d50}, \ref{table:validation_cec17_d100} summarize the results for dimensions 10, 30, 50 and 100, respectively. Inspecting these results we arrive at the following insights:
\begin{itemize}
\item EBOwithCMAR is the best algorithm for dimensions 10 and 50, whereas ELSHADE-SPACMA is the best algorithm for dimensions 50 and 100.
    
 \item Most of the new proposals (AEO, GSKA, and SSA) are statistically worse than competitive algorithms proposed back in 2017 (EBOwithCMAR) and the newest proposal ELSHADE-SPACMA.

\item For dimensions 10, 30, and 50, there are no significant differences among EBOwithCMAR, ELSHADE-SPACMA and jSO. However, for dimension 100 the recent algorithm ELSHADE-SPACMA is statistically better than them.
\end{itemize}

\begin{figure}[ht]
  \centering
\subfloat[Dimension 10]{
\includegraphics[width=0.45\textwidth]{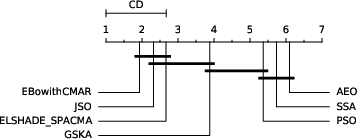}
}
\hfill
\subfloat[Dimension 30]{
  \includegraphics[width=0.45\textwidth]{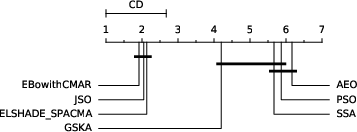}
}\\
\subfloat[Dimension 50]{
  \includegraphics[width=0.45\textwidth]{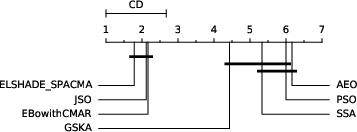}
}
\hfill
\subfloat[Dimension 100]{
  \includegraphics[width=0.45\textwidth]{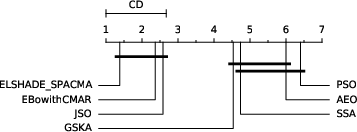}
}
\caption{\label{fig:CD}Critical distance plots for the CEC'2017 benchmark}
\end{figure}

As stated in Guideline \#2, more statistical tests beyond those used so far can be applied to reinforce our conclusions on the case study. We next discuss on the results of two new tests utilized for this end: the critical distance, and Bayesian tests.

Figure \ref{fig:CD} shows the statistical relevance of the differences between the average rankings over the CEC'2017 benchmark. The Critical Distance (CD) value (given by a Nemenyi post-hoc test at a significance $\alpha=0.05$) indicates the minimal absolute distance between two average rankings to be declared as statistically different to each other. It can be observed that among the new proposals under comparison, only ELSHADE-SPACMA is competitive with respect to the best performing approaches, though GSKA is also competitive for small problems. When the dimensionality increases, the distance between GSKA and the best algorithms (ELSHADE-SPACMA, EBOwithCMAR and jSO) increases, and the ranking difference with SSA is reduced. 

In dimension 10 the gaps between the ranks of the different algorithms become narrower. In order to arrive at more insightful conclusions, we apply several Bayesian tests over the results obtained for this dimensionality value, visualizing the adjusted Bayesian probability in barycentric coordinates. Bayesian analysis performed over the results of selected pairs of algorithms yields the probability that one solver outperforms another, based on the objective function values obtained by each of them over all runs and problems of the benchmark. The computed probability distribution displayed in barycentric coordinates after Monte Carlo sampling, depicting three regions: one where the first algorithm outperforms the second; a second one with the converse case (the second outperforms the first); and a region of practical equivalence where the results attained by each algorithm can be considered to be statistically equivalent to each other. To decide on this equivalence, a parameter called \emph{rope} indicates the minimum difference between the scores of both methods for
them to be considered significantly different to each other. 

At this point it is important to highlight the fact that \emph{rope} denotes a threshold imposed on the absolute difference of fitness values between the two compared algorithms. Consequently, \emph{rope} is interpretable and overcomes acknowledged issues identified around the use of p-values and significance levels in studies resorting to NHST for statistical assessment \cite{benavoli2017time}.

Turning back the focus on our case study, we consider several specific pairs of algorithms: (EBOwithCMAR, jSO), (jSO, GSKA) and (SSA, GKSA). We have considered only the first 20 runs for each function to make the processing time of the test computationally affordable. The \textit{rope} value is set to 20. Figure \ref{fig:bayesian_plots_cec17} depicts the Bayesian plots for each of the considered pairs, from which we confirm the following facts:
\begin{figure}[ht]
  \centering
  \subfloat[EBOwithCMAR vs jSO]{
    \includegraphics[width=0.4\textwidth]{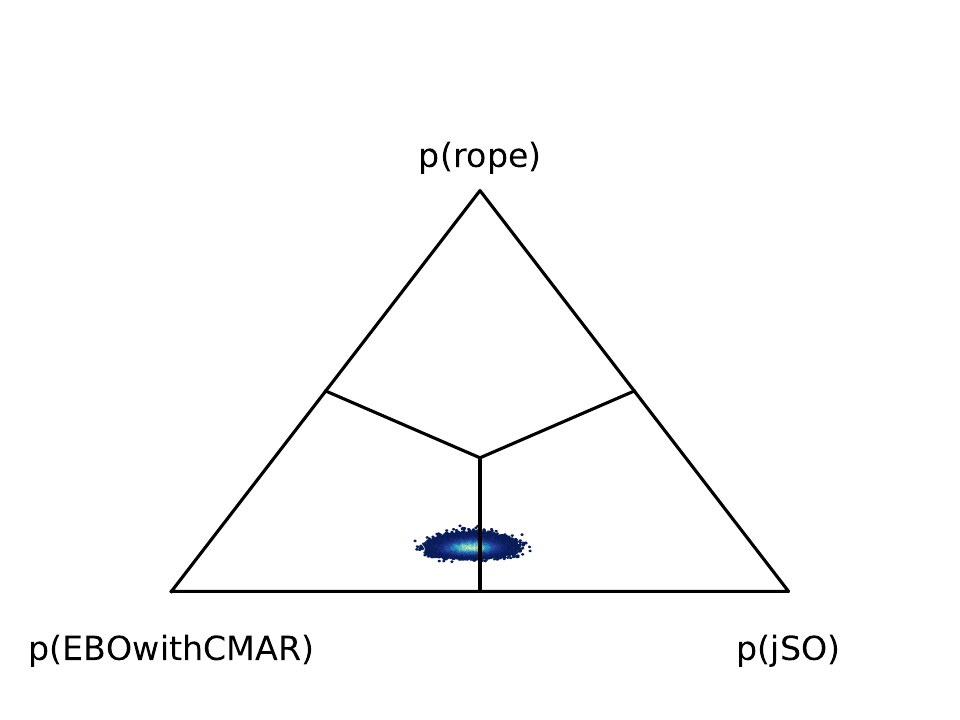}
  }
  \subfloat[SSA vs GSKA]{
    \includegraphics[width=0.4\textwidth]{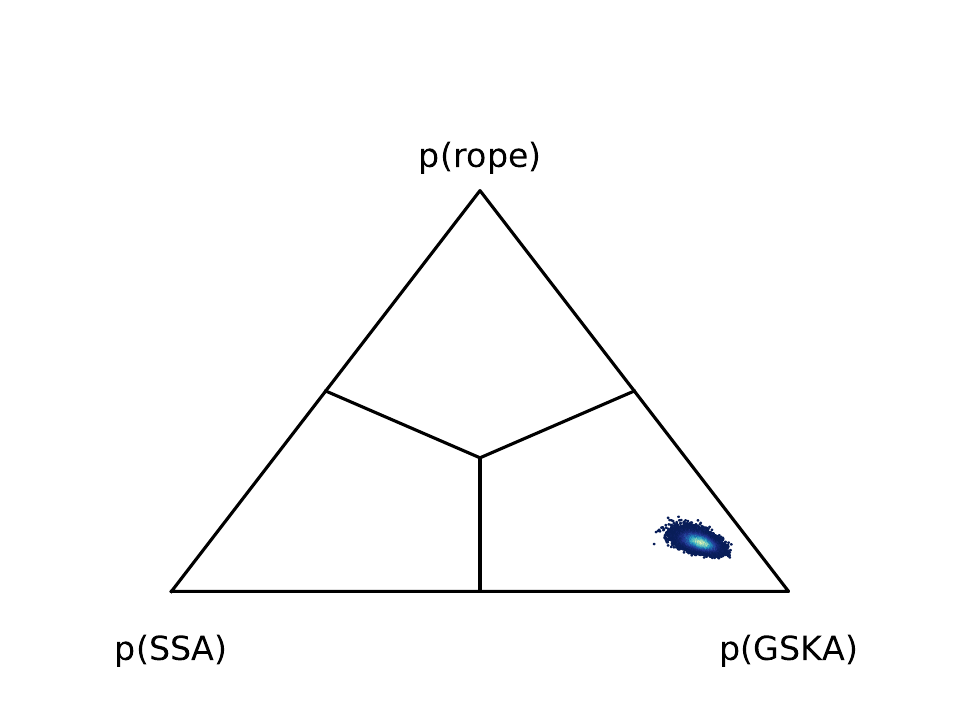}
  }\\
  \subfloat[jSO vs GSKA]{
    \includegraphics[width=0.4\textwidth]{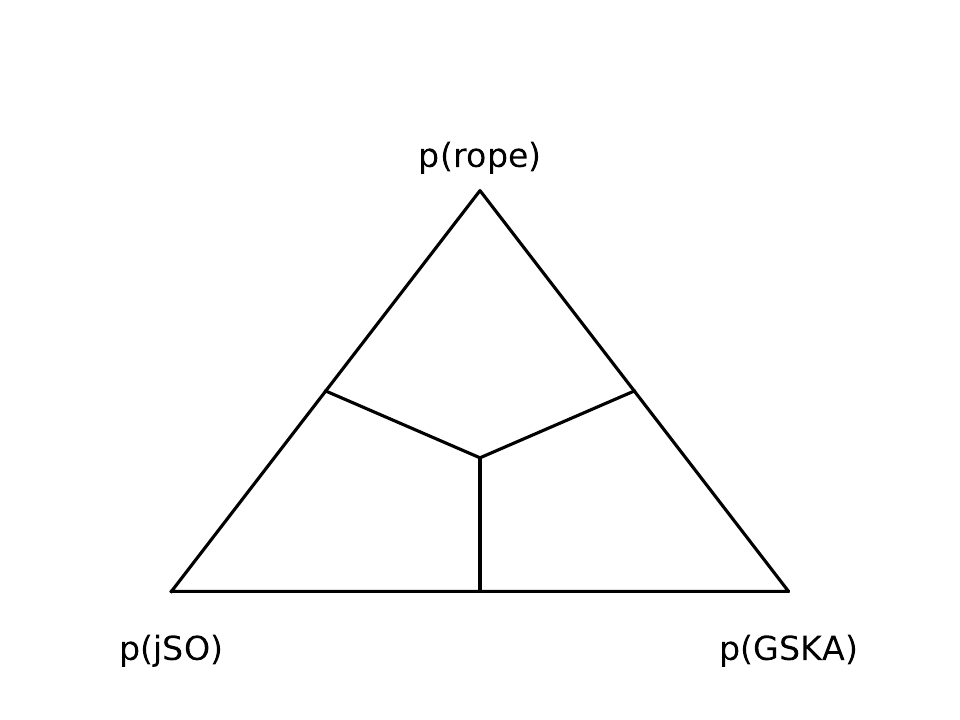}
  }
  \subfloat[jSO vs ELSHADE-SPACMA]{
    \includegraphics[width=0.4\textwidth]{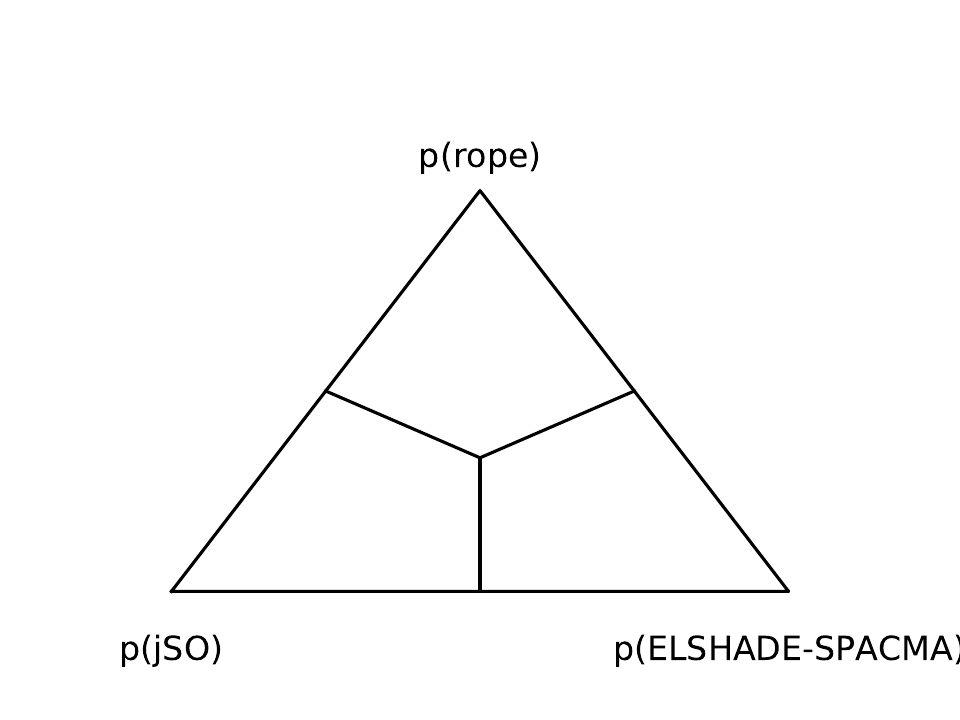}
  }
  \caption{Bayesian plots for the CEC'2017 benchmark and dimension 10.}
  \label{fig:bayesian_plots_cec17}
\end{figure}

\begin{itemize}[leftmargin=*]
\item There are almost no relevant differences among the results attained by ELSHADE-SPACMA, EBOwithCMAR and jSO.

\item For dimensions 30, 50, and 100 there are clearly two groups:
one made up by EBOwithCMAR, ELSHADE-SPACMA, and jSO; and another one, made up by GSKA, SSA, PSO, and AEO.

\item The lack of sampled point in the plots comparing jSO and GSKA reveals that the probability that GSKA achieves better results than jSO is exactly zero. This is specially relevant because the threshold value used is relatively high given the range of the objective functions characterizing the benchmark.

\item Differences between SSA and GKSA can be claimed to be significant for dimension 10, with GKSA emerging as the best performing approach.

\end{itemize}

Results show that ELSHADE-SPACMA outperform the previous winning algorithms EBOwithCMAR and jSO. Thus, it could be considered the new state-of-the-art method for this benchmark. In order to conduct a deeper analysis, our study includes a direct comparisons of ELSHADE-SPACMA against EBOwithCMAR and jSO. The results of this comparison are shown in Table \ref{table:ranking_comp_cec}, where it can be observed that among the most competitive algorithms, ELSHADE-SPACMA continues obtaining the best overall results. 

\begin{table}[ht]
  \centering
  \caption{Average ranking of the most competitive algorithms for each dimension}\label{table:ranking_comp_cec}
  \vspace{2mm}
  \begin{tabular}{lccccc}
    \toprule
    Algorithm   & D10                & D30            & D50   & D100 & Mean \\
    \midrule
    EBOwithCMAR     & \textbf{1.683}       & \textbf{1.817}       & 2.100          & 2.300          & 1.9750\\
    ELSHADE-SPACMA  & 2.267          & 2.133          & \textbf{1.783}       & \textbf{1.250}       & \textbf{1.858}\\
    jSO             & 2.050          & 2.050          & 2.117          & 2.450          & 2.1667\\
    \bottomrule
  \end{tabular}
\end{table}

\begin{table}[ht]
  \centering
  \caption{Results of the Wilcoxon test for the most competitive algorithms in the CEC'2017 benchmark for different dimension values.}
  \vspace{2mm}
  \label{table:wilcoxon_comp_cec17}
    \begin{tabular}{lclll}
      \toprule
      Algorithms & D10 & D30 & D50 & D100\\
      \midrule
      ELSHADE-SPACMA vs EBOwithCMAR & 0.027 & 0.226 & 0.509 & 3.128e-04\\
      EBOwithCMAR vs jSO & 0.225 &  0.431 & 0.054 & 0.477\\
      ELSHADE-SPACMA vs jSO & 0.796 & 0.706 & 0.556 & 2.254e-05\\
      \bottomrule
    \end{tabular}
\end{table}

Table \ref{table:wilcoxon_comp_cec17} reports the results of pairwise comparisons among ELSHADE-SPACMA, EBOwithCMAR and jSO with the Wilcoxon test. These results reveal that there are statistical differences only in dimension 100. This finding is confirmed in Table \ref{table:validation_comp_cec17_d100}, in which we provide the results corrected with the Holm method to account for the family-wise error in dimension 100.

\begin{table}[ht]
  \centering
  \caption{Statistical validation for the most competitive algorithms in the CEC'2017 benchmark and dimension 100 (ELSHADE-SPACMA is the control algorithm).}
  \vspace{2mm}
  \label{table:validation_comp_cec17_d100}
    \begin{tabular}{lcl}
      \toprule
      ELSHADE-SPACMA versus & Wilcoxon p-value & Wilcoxon p-value$*$\\
      \midrule
      jSO         &  2.254e-05 &  4.507e-05 ${\surd}$ \\
      EBOwithCMAR & 3.128e-04  &  3.128e-04 ${\surd}$ \\
      \bottomrule
      \noalign{\vskip 2mm}   \multicolumn{3}{l}{$\surd$: statistical differences exist with significance level $\alpha=0.05$.}\\
      \multicolumn{3}{l}{$^*$: p-value corrected with the Holm procedure.}\\&&
    \end{tabular}
\end{table}

\paragraph{COCO benchmark}

We continue our discussions on this first case study with the results over the COCO benchmark. To this end, we first use the source code and tools available at
\url{https://github.com/numbbo/coco} to run the experiments and obtain results of the
performance of every algorithm. Furthermore, these tools allow including in the benchmark the competitive algorithm proposed in 2009 for this benchmark.
\begin{figure}[ht]
  \centering
  \subfloat[Dimension 2]{
    \includegraphics[width=0.32\textwidth]{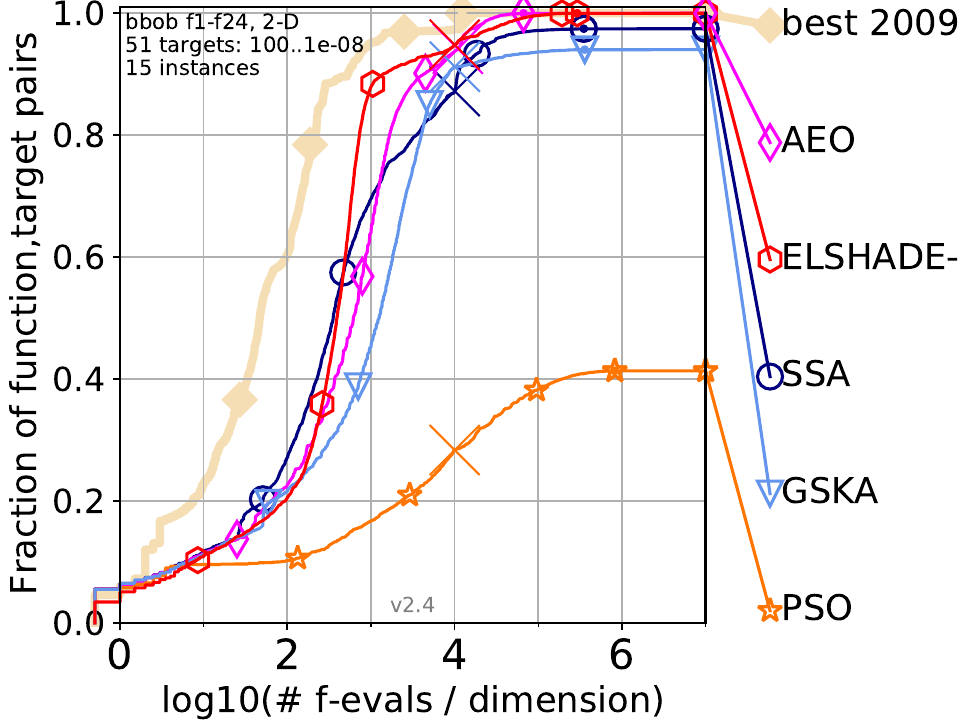}
  }
  \subfloat[Dimension 3]{
    \includegraphics[width=0.32\textwidth]{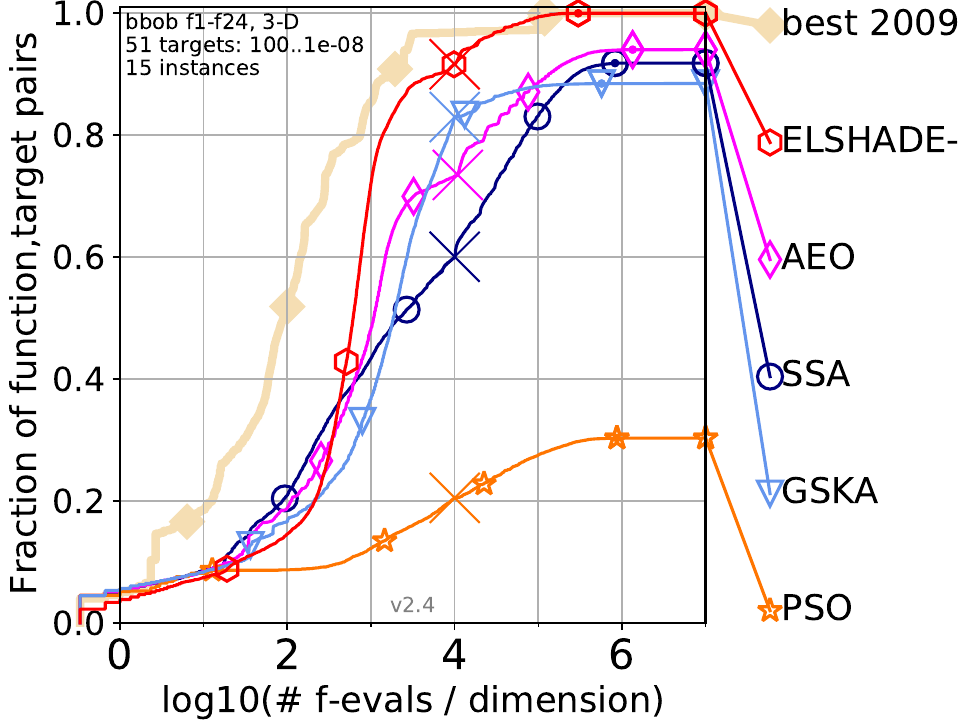}
  }
  \subfloat[Dimension 5]{
    \includegraphics[width=0.32\textwidth]{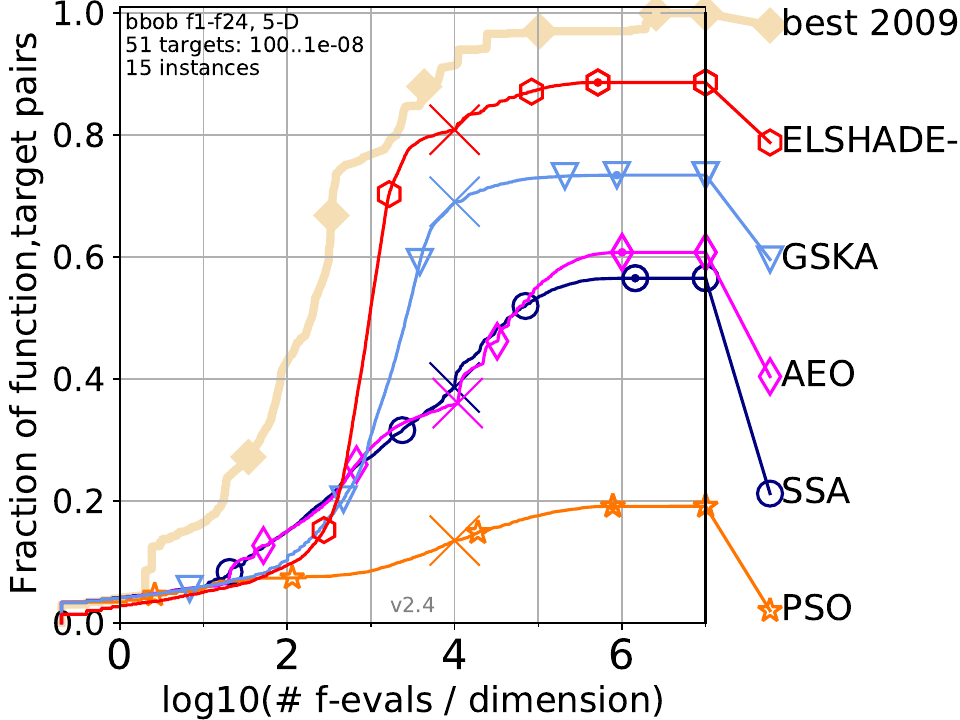}
  }\\
  \subfloat[Dimension 10]{
    \includegraphics[width=0.32\textwidth]{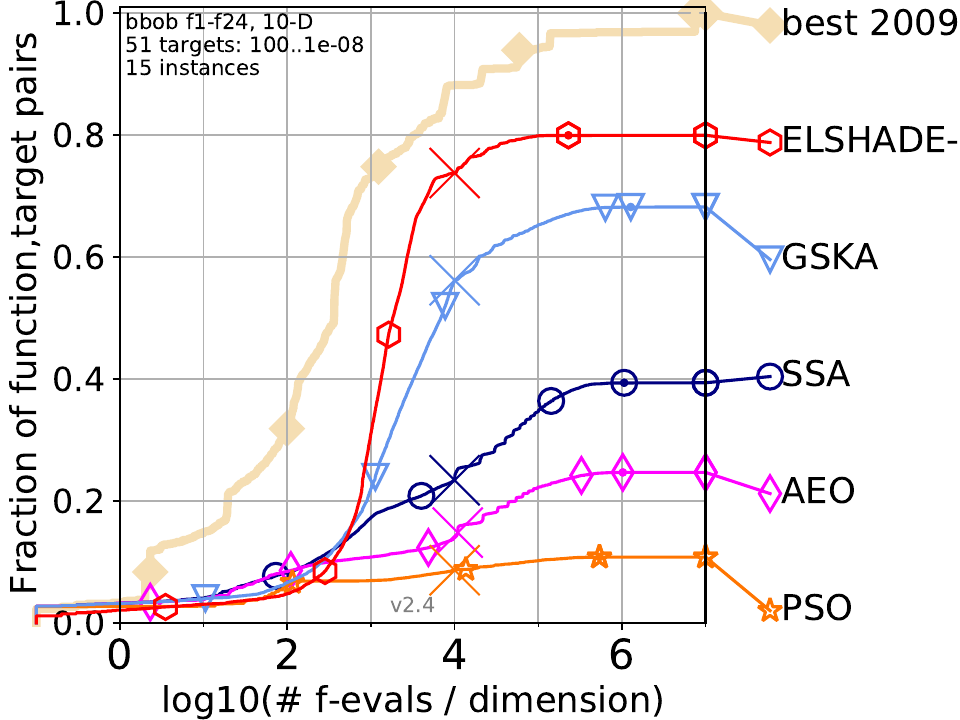}
  }
  \subfloat[Dimension 20]{
    \includegraphics[width=0.32\textwidth]{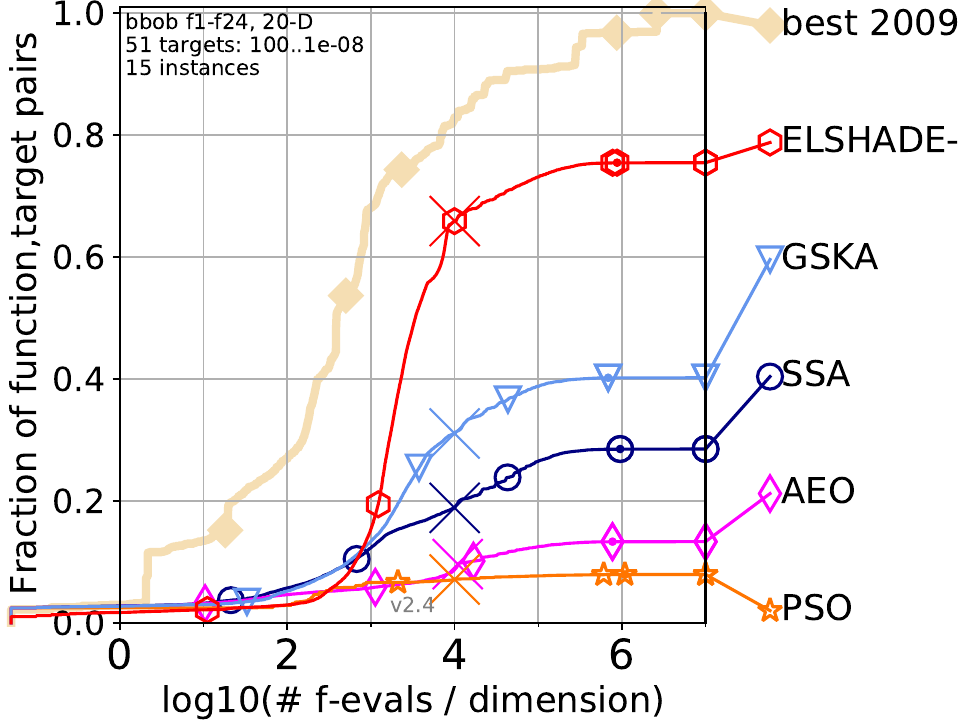}
  }
  \subfloat[Dimension 40]{
    \includegraphics[width=0.32\textwidth]{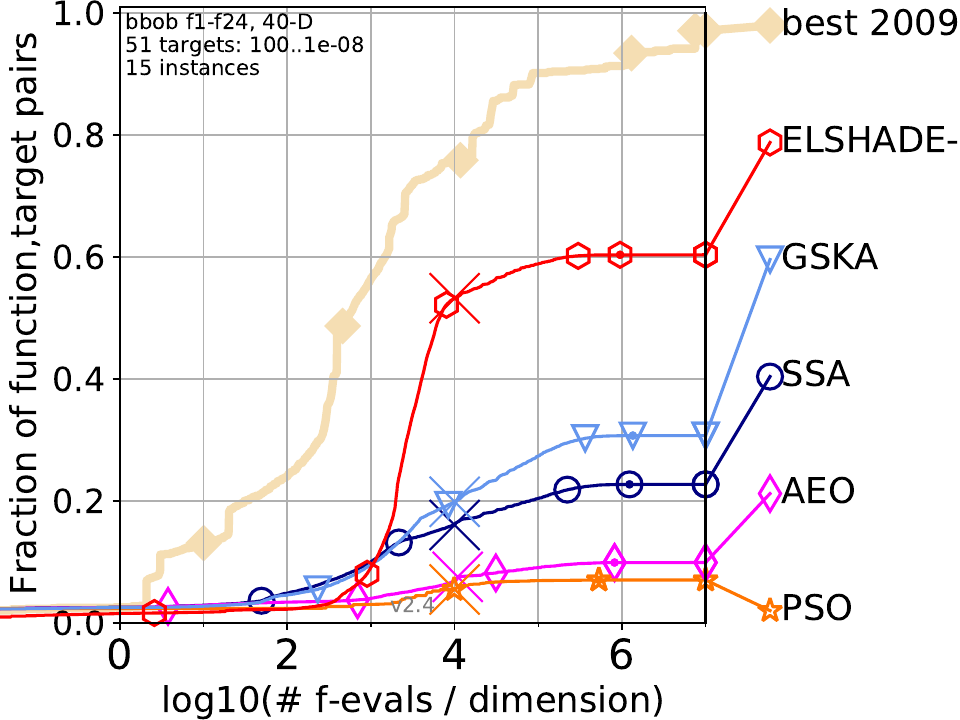}
  }
  \caption{Ratio of solved problems versus fitness evaluations for the COCO benchmark.}
  \label{fig:coco}
\end{figure}

Figure \ref{fig:coco} visualizes, for each algorithm, the ratio of solved problems (problems for which the error obtained is lower than a threshold) when the number of evaluations increases. The benchmark is designed for different dimension values, so Figure \ref{fig:coco} includes a subplot for each one of them. It can be observed that:
\begin{itemize}[leftmargin=*]
\item For low dimensionality values (e.g. 2 and 5), the results obtained for the other proposals are very similar and competitive, except for the reference algorithm (PSO). However, this behavior changes when the dimensionality increases.

\item For dimensionality equal to 5, ELSHADE-SPACMA (the acronym is shortened in Figure \ref{fig:coco}) and GSKA are notably better than SSA and AEO. For dimension 10, this noted improvement increases further.

\item For dimensionality values 20 and 40, GSKA still performs better than AEO, SSA and GSKA, but the difference reduces dramatically. GSKA exhibits a better behavior for lower dimensionality values.

\item For dimensionality values 20 and 40, ELSHADE-SPACMA is the algorithm that scores best when compared to the other proposals by a large gap. However, it does not approach the results of the best algorithm reported in 2009.
\end{itemize}

Following the same procedure as in the CEC'2017 benchmark, we have also considered whether to apply statistical tests. Unfortunately, the reduced number of runs for each function (15) is too low to allow for statistical tests with minimum significance guarantees.

\paragraph{Conclusions considering both benchmarks}

After analyzing together all experimental results discussed previously, our conclusions can be summarized as follows:
\begin{itemize}[leftmargin=*]

\item GSKA obtains better results than AEO and SSA.

\item For a lower budget in terms of number of objective function evaluations, GSKA is better than the other algorithms. On the contrary, when the number of evaluations is increased, EBOwithCMAR and jSO obtain better results.

\item The new proposals (AEO, GSKA and SSA) perform competitively when compared to classic optimization algorithms (e.g. PSO), but none of them can rival modern solvers like the standing winners of renowned competitions.

\item GSKA, for small dimensionality values, is better than other compared algorithms, but its advantage is reduced when the dimension increases.

\item ELSHADE-SPACMA has shown a rather opposite behavior: it improves as the dimensionality value increases. It is actually the only algorithm that is able to outperform existing competitive algorithms, especially for higher dimensionality values. For example, for dimension 100 it is even statistically better than the previous winning algorithms. In the COCO benchmark, on the other hand, although it is very competitive, it is not better than the previous winner of the 2009 competition. This could be due to the lower dimensionality values used in this competition.
\end{itemize}

All in all, the performance of recent meta-heuristic algorithms can be of interest when compared to that offered by classic approaches, but does not reach the levels of performance that are achieved by modern optimization methods.

\subsubsection{Components analysis and Tuning as per Guideline \#3}

Comparing with just reference and/or state-of-the-art algorithms is not enough. Following Guideline \#3 (Section \ref{sec:guideline3}), it is important for new proposals to analyze the behavior of the algorithm to gauge the contribution of each component. In this use case, since we have several new algorithms and the goal is to compare them fairly, analyzing the different components of each one of them because could entail an extensive study that goes beyond the scope of this work. For a example of component analysis, we refer to Section~\ref{sec:case2_guideline3}. 

Following Guideline \#3, the use of an automatic tuning mechanism is also recommended for the different algorithms in the comparison. In particular, we focus on the new proposal algorithms (AEO, GSKA, and SSA), because winners of the competitions, EBOwithCMAR and jSO, have been previously optimized by their authors for the competition. However, in our case the original publication where the most promising algorithm (GSKA and ELSHADE-SPACMA were proposed already considered the same benchmark (CEC'2017), so the parameter value used in this seminal reference are assumed to be the most adequate ones for this benchmark used in the case study. On the other hand, despite originally evaluated over a different benchmark, AEO has no parameters to be optimized. Its only free parameter (the population size) was analyzed in several works of its authors, concluding that \textit{varying the population size does not result in a significant change in precision} \cite{zhaoArtificialEcosystembasedOptimization2020}. Therefore, we center our discussion on the second-best performing algorithm in the results (SSA):

\paragraph{Tuning SSA}

To illustrate this process, we tune the two parameters driving the search behavior of the SSA approach, namely, the so-called \emph{swarm size} and \emph{attraction factor}, the lower value of the gliding distance $d_g$. Since there are only two parameters, instead of using a tool such as the ones commented in Subsection \ref{sec:guid4_parameter}, we have applied a grid search for dimensions 10, 30, and 50 (dimension 100 was omitted for the sake of computational affordability). In this grid search, different values are explored for each parameter:
\begin{itemize}[leftmargin=*]
\item For the swarm size, we have tested values \textbf{10}, \textbf{25}, \textbf{50} (recommended by authors), \textbf{75}, and \textbf{100}.
\item For the attraction factor, we have considered values equal to \textbf{0.1}, \textbf{0.3}, \textbf{0.5} (recommended by authors), \textbf{0.7}, \textbf{0.9}, and \textbf{1.0}.
\end{itemize}

\begin{table}
  \centering
  \subfloat[Dimension 10]{
  \begin{tabular}{cp{3em}rrrrr}
    \toprule
    & & \multicolumn{5}{c}{\textbf{swarm size}} \\
    &  & \multicolumn{1}{c}{\textbf{10}} & \multicolumn{1}{c}{\textbf{25}} & \multicolumn{1}{c}{\textbf{50}} & \multicolumn{1}{c}{\textbf{75}} & \multicolumn{1}{c}{\textbf{100}}\\
    \midrule
    \multirow{6}{*}{\rotatebox[origin=c]{90}{\textbf{attraction}}}
    & \textbf{0.1} & 20.40 & 18.05 & 17.82 & 17.35 & 17.52 \\
    & \textbf{0.3} & 19.87 & 15.58 & 17.02 & 14.05 & 13.58 \\
    & \textbf{0.5} & 19.30 & 18.38 & 13.88 & 12.65 & 10.25 \\
    & \textbf{0.7} & 17.80 & 17.25 & 13.08 & 12.55 & \textbf{8.22} \\
    & \textbf{0.9} & 18.87 & 18.15 & 14.75 & 10.82 & 10.35 \\
    & \textbf{1.0} & 18.83 & 18.78 & 14.82 & 13.48 & 11.55 \\
    \bottomrule
  \end{tabular}
}
\\
\vspace*{1em}
    \subfloat[Dimension 30]{
  \begin{tabular}{cp{3em}rrrrr}
    \toprule
    & & \multicolumn{5}{c}{\textbf{swarm size}} \\
    &  & \multicolumn{1}{c}{\textbf{10}} & \multicolumn{1}{c}{\textbf{25}} & \multicolumn{1}{c}{\textbf{50}} & \multicolumn{1}{c}{\textbf{75}} & \multicolumn{1}{c}{\textbf{100}}\\
    \midrule
    \multirow{6}{*}{\rotatebox[origin=c]{90}{\textbf{attraction}}}
    & \textbf{0.1} & 19.75 & 19.68 & 14.72 & 13.72 & 12.55 \\
    & \textbf{0.3} & 18.75 & 17.32 & 15.92 & 11.95 & 12.82 \\
    & \textbf{0.5} & 18.15 & 19.28 & 16.48 & \textbf{9.25} & 10.68 \\
    & \textbf{0.7} & 19.88 & 15.88 & 14.32 & 12.65 & \textbf{9.35} \\
    & \textbf{0.9} & 18.78 & 18.48 & 15.22 & 11.85 & 12.38 \\
    & \textbf{1.0} & 20.35 & 18.68 & 16.45 & 15.28 & 14.42 \\
    \bottomrule
  \end{tabular}
}
\\
\vspace*{1em}
\subfloat[Dimension 50]{
  \begin{tabular}{cp{3em}rrrrr}
    \toprule
    & & \multicolumn{5}{c}{\textbf{swarm size}} \\
    &  & \multicolumn{1}{c}{\textbf{10}} & \multicolumn{1}{c}{\textbf{25}} & \multicolumn{1}{c}{\textbf{50}} & \multicolumn{1}{c}{\textbf{75}} & \multicolumn{1}{c}{\textbf{100}}\\
    \midrule
    \multirow{6}{*}{\rotatebox[origin=c]{90}{\textbf{attraction}}}
    & \textbf{0.1}& 20.12 & 17.42 & 16.68 & 15.08 & 16.75 \\       
    & \textbf{0.3}& 18.58 & 15.82 & 12.02 & 10.75 & 13.12 \\       
    & \textbf{0.5}& 19.38 & 17.65 & 15.05 & 13.82 & 11.25 \\       
    & \textbf{0.7}& 17.55 & 16.48 & 12.62 & 13.25 & \textbf{10.28} \\      
    & \textbf{0.9}& 21.25 & 18.02 & 15.25 & 14.08 & 11.65 \\       
    & \textbf{1.0}& 18.05 & 19.92 & 16.68 & 12.88 & 13.55 \\
    \bottomrule
  \end{tabular}
}
\caption{Average ranking for different attraction and swarm size values for SSA algorithm and dimension}
\label{tab:tuning} 
\end{table}

As a result of this grid search, the $5\times 6 = 30$ parameter combinations have given rise to 30 configurations of SSA that we have compared to each other in terms of average rank. Such results are given in Table \ref{tab:tuning}, where we can notice that:
\begin{itemize}[leftmargin=*]
\item The best results are obtained with an attraction factor of 0.7 and swarm size equal to 100 for dimensionality values of 10 and 50. Furthermore, this configuration is the second best for dimension 30. Thus, these values can be declared to be the recommended setting for the algorithm.

\item In general, results are better with a larger swarm size, and with a medium-range value of the attraction factor.
  
\item The value recommended by the authors for the attraction value (0.5) is close to the best obtained during the tuning process (0.7). Moreover, it is the best one for dimension 30.
  
\item The recommended value for its authors in swarm size, 50, it is not adequate for this benchmark.
\end{itemize}

We have concluded that the net SSA results improve via parameter tuning. However, we must assess whether this performance improvement has any impact on the comparison previously discussed in Section \ref{sec:cec17_results}. As a global recommendation, parameters of all compared algorithms should be tuned. However, in this case the most competitive algorithms (ELSHADE-SPACMA, EBOwithCMAR, jSO and GSKA) were originally tuned over this benchmark. As a result, the only competitive approach that requires tuning is SSA. The rest of algorithms were not tuned considering its non-competitive results and the limited computational resources.

\begin{table}[ht]
  \centering
  \caption{Average ranking of the tuned algorithms for each dimension\label{table:ranking_cec_tuned}}
  \vspace{2mm}
  \begin{tabular}{llll}
    \toprule
    Algorithm      & D10            & D30            & D50            \\
    \midrule
    ELSHADE-SPACMA & 2.667          & 2.133          & \textbf{1.783} \\
    EBOwithCMAR    & \textbf{1.933}       & \textbf{1.917}       & 2.167          \\
    jSO            & 2.317          & 2.050          & 2.117          \\
    GSKA           & 3.883          & 4.200          & 4.433          \\
    PSO            & 5.400          & 5.933          & 6.067          \\
    SSA (Tuned)    & 5.533 (-0.40)  & 5.433 (-0.23)  & 5.200 (+0.13)  \\
    AEO            & 6.267          & 6.333          & 6.233          \\
    \bottomrule
  \end{tabular}
\end{table}

In Table \ref{table:ranking_cec_tuned} are shown the results of tuned algorithms, showing for the tuned algorithm (SSA) the difference in the average ranking in comparison with the non-tuned version. Dimension 100 was not included because there was no tuning performed in this dimension. It can be observed that the tuned version of the algorithm achieves an improvement over its non-tuned counterpart for dimensions 10 and 30, and it is worse for dimension 50. However, the improvement is not enough to change the relative position of the algorithm in the ranking.

\subsubsection{Justifying the usefulness of the algorithm as per Guideline \#4}

Following Guideline \#4 (Section \ref{sec:guideline4}), there are several perspectives from which one may claim the usefulness of an algorithm:
\begin{itemize}[leftmargin=*]
\item \emph{Quality of the results}: in this case, the majority of the evaluated algorithms have been found to be not competitive enough in comparison with the \textit{state-of-the-art} methods. Only ELSHADE-SPACMA has been able to improve them, thus becoming the current \textit{state-of-the-art} algorithm in the benchmark considering these results. Also, the comparisons between the different algorithms expose significant differences between them. Therefore, the results can be considered to be interesting and informative for researchers aiming to elaborate further on their design. Furthermore, while GSKA is not competitive against jSO and EBOwithCMAR, it is able to perform best when the number of of function evaluations is low. In many real-world problems, the evaluation of a solution can be very costly in terms of computational resources (e.g. when the fitness value is produced by long computer simulations). Under these circumstances, it is essential to rely on algorithms capable of obtaining good results within a small number of evaluations.

\item \emph{Technical novelty}: the compared proposals not only have very different biological inspiration, but they differ notably regarding their algorithmic behavior. In particular, ELSHADE-SPACMA, SSA and GSKA propose interesting ideas that could be considered for new algorithms. ELSHADE-SPACMA combines different previous algorithms (SHADE-ILS and CMA-ES) with a novel mechanism to improve the diversity in the population, a mutation operator that not only considers the best solutions but also the worst ones. Another contribution is the adaptation of one parameter used to increase diversity in the early stages of the search. With these two main changes that make it different than the previous LSHADE-SPACMA algorithm, ELSHADE-SPACMA is able to improve the results of the competitive EBOwithCMAR and jSO algorithms. LSHADE-SPACMA failed to do so. On the other hand, SSA exploits the idea of ranking the different solutions to create several categories (the current best, the best ones, and the normal, which is the largest group), and use a different mutation method considering the category of each individual. In addition, solutions which cannot get improved are periodically restarted. These two strategies yield a very straightforward yet effective optimization algorithm. GSKA also suggests other interesting concepts from the technical point of view: it updates the variables of each individual under two possible criteria, among which the selected one is updated during the search to increase the exploitation during the run of the algorithm. Furthermore, in the exploration GSKA uses for each individual the most similar ones in fitness, the immediate better one and the immediate worse one.

\item \emph{Methodological contribution}: There is no methodological contribution.

\item A special attention should be paid to the \emph{simplicity}, in which SSA outstand in the benchmark. However, modifications should be applied to improve its results and avoid its apparently premature convergence.
\end{itemize}

\subsubsection{Summary of the use case}

To conclude, this first use case follows most of the guidelines of our proposed methodology. The main procedures followed in the use case are highlighted in Figure \ref{fig:case_checklist}. In addition, we briefly describe now the main actions taken for each of the proposed guidelines:

\begin{figure}[ht]
  \centering
  \includegraphics[width=\textwidth]{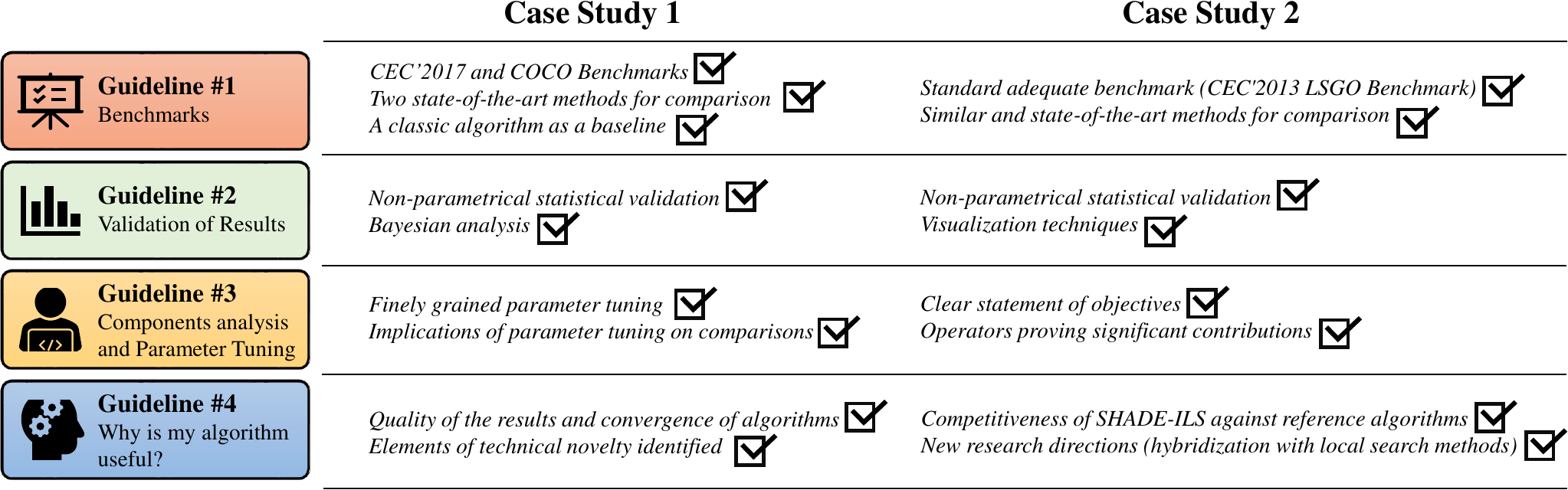}
  \caption{Checklist of the guidelines' recommendations followed by the use cases.}
  \label{fig:case_checklist} 
\end{figure}

\begin{itemize}[leftmargin=*]
\item Guideline \#1: We have selected two different real-parameter benchmarks, CEC'2017 and COCO. Both are widely accepted by the community working in real-parameter optimization. We have also compared our algorithms against competitive solvers that have won competitions using the same benchmarks, as well as a reference baseline for swarm intelligence (i.e. PSO).
    
\item Guideline \#2: following this guideline, we have shown and validated the results according to the good practices described in Section \ref{sec:guideline2}, including statistical non-parametric tests and Bayesian analysis. Furthermore, we have provide evidence on how the identification of the best algorithm can strongly be biased by the stopping criterion in use (e.g. the maximum number of objective function evaluations). 
    
\item Guideline \#3: following the suggestions in this guideline, we have applied a simple tuning process and unveiled that the right parameter values can influence the results and conclusions held from the previous comparison.

\item Guideline \#4: in this use case, the comparisons between the different algorithms could be justified as per the relevant differences found among them. Only one of the compared algorithms (ELSHADE-SPACMA) has been able to outperform other existing competitive algorithms. Moreover, one of them (GSKA) has shown a superior behavior when the number of function evaluations is low. Nevertheless, the fact that the majority of these modern algorithms could not improve previous competitive solvers is a relevant fact that should stimulate fairer comparison studies in prospective works with new and/or improved versions of bio-inspired optimization algorithms. Finally, some of these algorithms pose innovative algorithmic ideas that should be investigated further towards their use in the design of new search methods.
\end{itemize}

\subsection{SHADE-ILS for large-scale global optimization}\label{sec:case_study_lsgo}

In this second case study we simulate the situation in which we design a new algorithm, SHADE-ILS, specially designed for large-scale global optimization. In this section we follows the guidelines for properly conducting the experiments, comparisons with other reference algorithms, and the analysis to put in value the advantages of our methodological proposal.

\subsubsection{Selecting the benchmark as per Guideline \#1}
\label{sec:case_guideline1_benchmark}

First, we have to choose the right benchmark for the experimental assessment of the performance of our newly proposed algorithm. Following the recommendations of Guideline \#1 (described in Section \ref{sec:guideline1}), we have to:
\begin{itemize}[leftmargin=*]
\item Properly select the benchmark: without any unexpected bias, with the right level of complexity, and for the type of problem addressed by the algorithm.

\item Enforce the usage of a standard benchmark that fulfills the previous requirements.
\end{itemize}

The selection of the benchmark cannot be done without considering the proposed algorithm, since it depends on the characteristics of the problem for which the algorithm was implemented (or the type of problems for which we want to test it). In our example, we have designed SHADE-ILS \cite{Molina2018}, an algorithm specially devised for real-parameter optimization problems that comprise a high number of variables. This family of optimization problems is collectively referred to as large-scale global optimization, for which several benchmarks have been proposed \cite{Tang2010,Lozano2011,Li2013a}. If any of them allows for an unbiased comparison, we should use it, avoiding in this way the design of our own benchmark. In particular, our first option is the CEC'2013 benchmark \cite{Li2013a}, since it is both the most recent and the most popular competition to date. Furthermore, its popularity yields many previous results that we can use for comparison purposes. Nevertheless, before proceeding further we have to verify whether the selected benchmark allows for good comparisons. For this purpose, information and data available about the benchmark should comply with several requirements:
\begin{todolist}[leftmargin=*]
\item[\done] \textit{Clear experimental conditions}: the experimental setup is well defined, and conditions are set the same for all algorithms.
\item[\done] \textit{The implementation of the benchmark is openly available}: the CEC'2013 benchmark is specially appropriate in this regard, as implementations of the problems comprising the benchmark are not only made publicly available, but also in several programming languages: C/C++, Matlab, Java, and Python\footnote{Code for the CEC'2013 Large Scale Global Optimization benchmark: \url{https://www.tflsgo.org/special_sessions/wcci2020.html\#new-code} (accessed on April 16th, 2020).}.
\item[\done] \textit{The optima is not at the center of the domain search}: all functions in the chosen benchmark are shifted to guarantee this feature.
\item[\wontfix] \textit{Functions are rotated}: Although this feature is not present in the chosen benchmark, the importance of this requirement is not as critical as the aforementioned shifting.
\item[\done] \textit{Presence of local optima}: In the benchmark there are several functions with different local optima. Actually, it is not the only criterion to provide functions with varying levels of difficulty. In particular, in this benchmark there are different degrees of interrelation between variables, which makes sense given the large dimensionality of the problems.
\end{todolist}

Summarizing, the above analysis concludes that the CEC'2013 benchmark for large-scale global optimization follows most of the requirements imposed by our methodology. This is the reason why we select it as benchmark for the experimentation.

\subsubsection{Selecting the performance measure as per Guideline \#1} \label{sec:case2_guideline1_performance}

Another important decision to make is the choice of an adequate performance measure. In this regard, we can measure not only the final fitness error (deviation with respect to the global optimum that is known \textit{a priori}), but also the error for different number of fitness evaluations (called the \emph{accuracy level}). This way, we can fairly measure the efficiency of the algorithms. There are two possibilities in this matter: i) to report the performance for each accuracy level; and ii) to provide the performance for the maximum number of fitness evaluations considered in the experiments. To show the results concisely, we will only discuss on the latter of these alternatives (i.e. the results for the maximum number of fitness evaluations). However, the study should be done in a similar fashion for each level of accuracy.

About the performance, there are also several possibilities: we can report the fitness error function by function, or we can compute an aggregate measure of performance (such as an average). Initially, we opt for an aggregate measure, considering two options:
\begin{itemize}[leftmargin=*]
\item \emph{Average ranking}, which is calculated by sorting the algorithms for each function based on its error (lower position to best ones). Then the average ranking is calculated so that an algorithm with a lower average ranking value is declared to perform better, on average, than other with a higher ranking value.
  
\item \emph{A particular measure proposed in the considered benchmark}, which assigns for each function a different score to each algorithm, based on its ranking position.
\end{itemize}

The performance measure recommended in competitions with the CEC'2013 benchmark is the second one of these options. However, we will first depict the average ranking, since it evaluates the performance of algorithms in a more general and understandable way.

\subsubsection{Selecting the reference algorithms as per Guideline \#1}
\label{sec:case2_guideline1_reference}

In order to do a right comparison, a clear criterion is needed to select the algorithms included in the comparison, aiming at fairly proving the convenience of the algorithm in regards to its competitive performance with other methods. Following the guidelines, we should:
\begin{itemize}[leftmargin=*]
\item \textit{Compare against reference algorithms}: in this benchmark DECCG \cite{Li2013a} will take this role.
  
\item \textit{Compare against similar algorithms}: this aspect is specially relevant when the proposed algorithm is a modified version of a previously published approach. In our case, SHADE-ILS can be deemed a new algorithm. However, other proposals featuring similar concepts were previously proposed in the literature, such as IHDELS \cite{molina2015iterative}. Following our guidelines, we have included these previous methods for their comparison to our proposal.

\item \textit{Compare against competitive algorithms}: this is often a hard decision to make, since it is difficult to scrutinize the entire state-of-the-art related to the optimization problem/algorithm/benchmark under consideration. However, since the benchmark is widely used in international competitions, we can use the winning approaches in these competitions as competitive algorithms to which to compare our proposed approach. As such, one of the solvers in this field is MOS-CEC2013 \cite{LaTorre2013c}, which has been the best algorithm in these competitions for years. Additionally, we include MLSHADE-SPA \cite{hadiLSHADESPAMemeticFramework2019} in our comparison, as it was reported to outperform MOS-CEC2013 results in the 2018 competition. Nowadays, there are other competitive algorithms, but we focus on the algorithms proposed until 2018, the year in which the algorithm was presented \cite{Molina2018}.
\end{itemize}

On balance, we compare our method against a considered previous version (IHDELS), a competitive algorithm (MOS-CEC2013), and a reference algorithm (DECCG).

\subsubsection{Testing and validating the results as per Guideline \#2} \label{sec:case_guideline2}

After the design of the experimentation, experiments are carried out, and results are validated. Following the recommendations about statistical validation in Guidelines \ref{sec:guideline1_ref_algs} and \ref{sec:guideline2_stat}, normality or homocedasticity tests should be performed. However, it has been proven that such tests do not usually pass for a benchmark as the chosen one \cite{garcia08_study_use_non_param_tests}. Therefore, we have opted for non-parametric tests, given that it is unlikely that normality and homocedasticity hold for the CEC'2013 benchmark.

Thus, the first step to take is to calculate the average ranking, followed by the non-parametric hypothesis test. In order to compare the algorithms, we resort to Tacolab \cite{tacolab}\footnote{Tacolab website: \url{https://tacolab.org/}}, a web tool that eases the application of different comparison methods among algorithms.
\begin{table}[ht]
\centering
\caption{Average ranking of the algorithms considered for the statistical comparison} \label{table:ranking}
\vspace{2mm}
\begin{tabular}{lc}
\toprule
Algorithm   & Ranking \\
\midrule
SHADE-ILS   & 1.967    \\
MLSHADE-SPA   & 2.433    \\
MOS-CEC2013 & 2.700    \\
  IHDELS      & 3.633    \\
  DECCG       & 4.267    \\
\bottomrule
\end{tabular}
\end{table}

Table \ref{table:ranking} shows the average ranking of the four algorithms under comparison over the CEC'2013 LSGO benchmark. We recall that SHADE-ILS is the new algorithmic proposal, whereas MLSHADE-SPA is another proposal presented in the same competition than IHDELS, and MOS-CEC2013 and DECCG are the two state-of-the-art methods considered as reference algorithms. The table depicts the average ranking computed from the relative position of the four methods when ranked for each of the functions in the benchmark. As can be observed in this table, SHADE-ILS exhibits a slightly better performance than MLSHADE-SPA and MOS-CEC2013, and a much better rank than IHDELS and DECCG. In particular, although the preceding approach (IHDELS) performed worse than MOS-CEC2013, SHADE-ILS renders a significantly better performance. This aspect is quite important, because it is not common to directly design a competitive algorithm from scratch.

As stated in Guideline \#2 (Section \ref{sec:guideline2}), performance measures like the average ranking are not conclusive, since performance gaps may occur due to the stochastic nature of the algorithms under comparison. This is the reason why these results should be further analyzed for elucidating whether the differences are significant. For this purpose, we use the Friedman rank-sum test. The p-value reported by this test is 4.87e-03, which is clearly significant at the $\alpha=0.05$ confidence level. Now that the aforementioned differences have been assessed, we can proceed towards the multiple comparison, including a familywise error rate correction tackled with the Holm procedure.
\begin{table}[ht]
  \centering
  \caption{Statistical validation (SHADE-ILS is the control algorithm)}
  \vspace{2mm}
  \label{table:validation}
  \begin{tabular}{lcl}
    \toprule
    SHADE-ILS versus & Wilcoxon p-value & Wilcoxon p-value$^*$ \\
    \midrule
    MLSHADE-SPA   & 1.51e-01 & 1.51e-01 ${\approx}$\\
    MOS-CEC2013   & 4.79e-02 & 9.58e-02 ${\approx}$\\
    IHDELS        & 1.53e-03 & 2.50e-02 ${\surd}$\\
    DECCG         & 8.36e-03 & 6.10e-03 ${\surd}$\\
    \bottomrule
    \noalign{\vskip 2mm}   \multicolumn{3}{l}{$\surd$: statistical differences exist with significance level $\alpha=0.05$.}\\
    \multicolumn{3}{l}{$^*$: p-value corrected with the Holm procedure.}\\
  \end{tabular}
\end{table}

Table \ref{table:validation} presents the results of this analysis. As can be observed, differences are significant between SHADE-ILS, DECCG and IHDELS. Regarding MLSHADE-SPA, there are not statistical differences with respect to SHADE-ILS. Finally, in the case of MOS-CEC2013, there are also no statistical differences after applying the Holm correction procedure and using a confidence level of 5\%, yet increasing the confidence level up to 10\% would make it possible to conclude that differences are statistically significant. This comparison should also consider different checkpoints, e.g., $1\%$, $10\%$ and $100\%$, or every $10\%$ of the maximum number of fitness evaluations available. This complementary analysis would reflect not only the final result of the algorithms, but also their convergence speed.
\begin{figure}[ht]
  \centering
  \includegraphics[width=.75\textwidth]{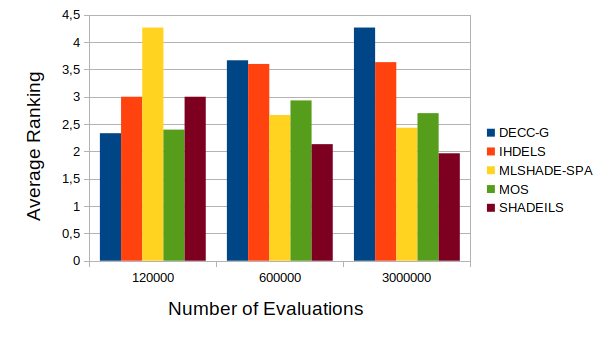}
  \caption{Average ranking of the algorithms for different numbers of fitness evaluations.}\label{fig:visual}
\end{figure}

Besides that, we indicated in Section \ref{sec:guidelines2_visualization} that a graphical visualization is useful for the analysis. In this case, we complement the study summarized in Table \ref{table:ranking} with Figure \ref{fig:visual}. In this figure, it is more evident that the differences between algorithms increase with the number of fitness evaluations: whereas MOS is better than DECCG and IHDELS, both MLSHADE-SPA and SHADE-ILS improve their results since 600,000 evaluations, showing that SHADE-ILS achieves the best average ranking with 600,000 and 3,000,000 evaluations. Ideally, a convergence plot could be more informative, but in the CEC'2013 benchmark the milestones posed by the competition are very reduced, so a bar plot like the depicted one results to be more helpful for the purpose of this analysis.

\subsubsection{Components analysis and tuning as per Guideline \#3}
\label{sec:case2_guideline3}

As mentioned before, a comparison of a proposed method with just reference and/or state-of-the-art algorithms is usually not enough. Following Guideline \#3 (Section \ref{sec:guideline3}), when analyzing the algorithm it is also important to clarify the objectives for the proposed design, and then show quantitative evidence of the claims about the behavior of the algorithm. This way, the study can shed light on the influence of the different components over the reported final results. In our use case, we do not explain the objectives and main ideas of the algorithm. Instead, we remark that the main changes featured by SHADE-ILS with respect to IHDELS is i) a modification of the Differential Evolution component (from SaDE to SHADE); and ii) the restart mechanism. We refer interested readers to \cite{Molina2018a} for further details.
\begin{table*}[ht!]
\vspace{-2mm}
  \centering
  \caption{Comparisons between different components of the proposal}
  \vspace{2mm}
  \input{comp_previous}

  \label{tab:comp_analysis}
\end{table*}

Table \ref{tab:comp_analysis} shows the results obtained by the different components of the algorithm. This table clearly exposes that the outperforming behavior of the proposed method is due to all its novel contributions, rather than a subset of them. Furthermore, these changes do not add complexity to the overall search process.

Following Guideline \#3, the use of an automatic tuning mechanism is also recommended for the different algorithms in the comparison. However, in our case we use the results reported by their authors in the contributions where the algorithms were first presented, so it is expected that these results were obtained by using the best parameter values. Regarding the parameter values of the SHADE-ILS proposal, they should be obtained by a tuning process, ideally conducted by an automatic tool. In our case, a manual tuning has been conducted due to computational constraints (in particular, processing time). If more resources for computation were available, a complete tuning process could be conducted by resorting to available tools such as the ones commented in Subsection \ref{sec:guid4_parameter}.

\subsubsection{Justifying the usefulness of the algorithm as per Guideline \#4}
\label{sec:case_guideline4}

Following Guideline \#4 (Section \ref{sec:guideline4}), there are several ways to show the usefulness of an algorithm:
\begin{itemize}[leftmargin=*]
\item \emph{Quality of the results}: in this case, given the good results in the comparisons against \textit{state-of-the-art} and reference methods, the scientific value of the contribution is clear. 

\item \emph{Technical novelty}: the combination of local search methods and the hybridization of SHADE are novel. However, for the sake of conciseness we will not elaborate here on the originality of these ingredients, as it would require an exhaustive review of the recent history of DE approaches for large-scale global optimization. We defer the reader to the analysis made in \cite{Molina2018} in this regard.

\item \emph{Methodological contribution}: SHADE-ILS improves a previous hybridization of DE with local search \cite{molina2015iterative} by embracing, among other algorithmic additions, Success-History based Adaptive Differential Evolution (SHADE) at its core, which culminates a historical series of adaptive DE solvers. This poses no doubt on the scientific contribution of this study, as can stimulate new research directions towards considering new local search methods hybridized with SHADE. 

\item A special attention should be given to the \emph{simplicity} of SHADE-ILS. In this algorithm the model is not very complex, and the number of parameters is simpler than other proposals (due that its components require few parameters). Besides, changes made with respect to IHDELS do not increase its number of parameters.
\end{itemize}

\subsubsection{Summary of the use case}

On a closing note, the use case depicted in this section follows most of the guidelines of our proposed methodology. As in the previous use case, the main procedures followed are highlighted in Figure \ref{fig:case_checklist}. In addition, we briefly describe now the main actions taken for each of the proposed guidelines:

\begin{itemize}[leftmargin=*]
\item Guideline \#1: we have resorted to the standard CEC'2013 benchmark, which is widely accepted by the community working on large-scale global optimization. Also, we have checked that the benchmark follows several of the requirements imposed by the guidelines. Finally, we have compared our proposed method against similar state-of-the-art techniques and a reference baseline from the field.
    
\item Guideline \#2: as has been shown throughout the discussion, the validation of the results has been done according to the good practices prescribed in Section \ref{sec:guideline2}, including non-parametric hypothesis tests. Also, we have also shown how several results can be properly visualized to make the outcome of the comparisons more understandable to the audience.
    
\item Guideline \#3: we have clearly highlighted the objectives of the algorithms, and we have compared the influence of the different novel elements of the proposed algorithm. Thus, we have shown that the good results are not influenced by just one component, but to the synergy between the different elements. We have also shown that the proposal is not unnecessarily complex. Finally, a tuning process has been also applied.

\item Guideline \#4: in this use case, the proposal of our method is easy to justify. SHADE-ILS not only improves a previous hybridization of DE with local search \cite{molina2015iterative}, but also surpasses MOS-CEC2013 and MLSHADE-SPA (albeit without statistical significance), which have dominated the competition over the last few years. This poses no doubt on the scientific contribution of this study, as can stimulate new research directions towards considering new local search methods hybridized with SHADE. In addition, the previous state-of-the-art algorithm, MOS, has been clearly surprised by SHADE-ILS and MLSHADE-SPA, hence becoming the most competitive algorithms (with a preference by SHADE-ILS as per its better performance and simplicity).
  
 \end{itemize}

\section{Conclusions and Outlook} \label{sec:conclusions}

\begin{figure}[ht!]
    \centering
    \includegraphics[width=\columnwidth]{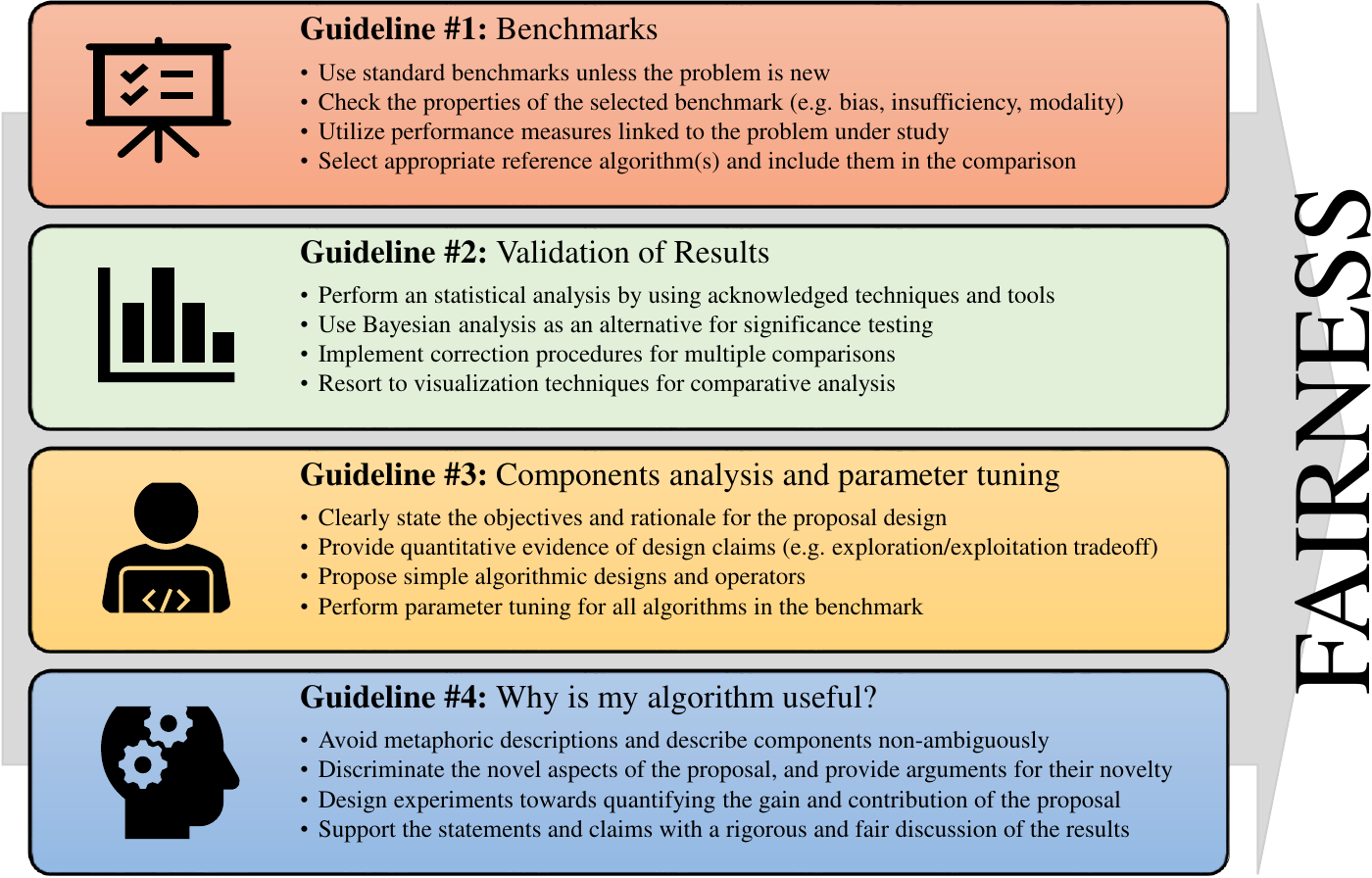}
    \caption{Guidelines composing the methodological framework for comparing meta-heuristics proposed in this work.}
    \label{fig:methodology}
\end{figure}

In this work we have stressed on the need for circumventing common mistakes and flaws observed in the field of bio-inspired optimization, particularly when new meta-heuristic algorithms are proposed and experimentally validated over benchmarks designed to this end. Specifically, we have reviewed and critically analyzed contributions dealing with experimental recommendations and practices related to meta-heuristics. Following our literature study, we have prescribed a set of methodological recommendations for preparing a successful proposal of bio-inspired meta-heuristic algorithms, from the definition of the experimentation to the presentation of the results. A number of useful techniques (graphically summarized in Figure \ref{fig:methodology}) have been suggested for prospective studies to implement our proposed methodological framework, in an attempt at ensuring fairness, coherence and soundness in future studies on the topic. Two different case studies have been designed to exemplify the application of our prescribed methodology, discussing on the results of the application of each guideline. Although both case studies deal with well-known benchmarks, we envision that our methodology can be a core part of the design process of meta-heuristic algorithms for real-world optimization problems, following the guidelines of our recently published tutorial in this matter \cite{osaba2021tutorial}. In those cases, the statistical validation of the results should not be considered the final step of the analysis: the significance of the results should be analyzed from different perspectives, and taking into consideration other measures of practical relevance (e.g. memory consumption).

In such a vibrant field, with new algorithmic proposals flourishing vigorously, common methodological grounds are urgently needed. Scientific advancements in years to come will only be achieved if the community reaches an agreement on how algorithms should be tested and compared to each other. This is indeed the aim of our work: to gather and group recommended practices around an unified set of systematic methodological guidelines. We sincerely hope that the material and prescriptions given herein will guide newcomers in their arrival to this exciting research avenue.

\section*{Acknowledgments}

This work was supported by grants from the Spanish Ministry of Science (TIN2016-8113-R, TIN2017-89517-P and TIN2017-83132-C2-2-R) and Universidad Polit{\'e}cnica de Madrid (PINV-18-XEOGHQ-19-4QTEBP). Eneko Osaba and Javier Del Ser would also like to thank the Basque Government for its funding support through the ELKARTEK and EMAITEK programs.  Javier Del Ser receives funding support from the Consolidated Research Group MATHMODE (IT1294-19) granted by the Department of Education of the Basque Government.

\small

\bibliographystyle{elsarticle-num}
\bibliography{ecj_bio_guidelines,revisores}

\end{document}

%% file: results_evol_cec_d10.tex
\subfloat[1-30\% evaluations]{
\begin{tabular}{l*{7}{S}}
\toprule
Algorithm &       \multicolumn{1}{c}{1}   &       \multicolumn{1}{c}{2}   &       \multicolumn{1}{c}{3}   &       \multicolumn{1}{c}{5}   &       \multicolumn{1}{c}{10}  &       \multicolumn{1}{c}{20}  &       \multicolumn{1}{c}{30}\\
  \midrule
AEO            &  \bfseries 1.700000 &  \bfseries 1.966667 &  2.600000 &  3.800000 &  4.966667 &  5.466667 &  5.800000 \\ 
EBOwithCMAR    &  4.500000 &  4.500000 &  4.033333 &  3.133333 & \bfseries 2.300000 &  \bfseries 1.900000 &  2.050000 \\
ELSHADE-SPACMA &  5.866667 &  5.600000 &  5.200000 &  4.133333 &  3.233333 &  2.266667 &  \bfseries 1.750000 \\
GSKA           &  2.833333 &  2.200000 &  \bfseries 2.200000 & \bfseries 2.366667 &  2.533333 &  3.266667 &  3.700000 \\ 
PSO            &  6.200000 &  6.400000 &  6.466667 &  6.466667 &  6.266667 &  6.166667 &  5.933333 \\ 
SSA            &  1.866667 &  3.166667 &  3.966667 &  4.666667 &  5.200000 &  5.433333 &  5.466667 \\ 
jSO            &  5.033333 &  4.166667 &  3.533333 &  3.433333 &  3.500000 &  3.500000 &  3.300000 \\ 
  \bottomrule
\end{tabular}
}
\\
\subfloat[40-100\% evaluations]{
\begin{tabular}{l*{7}{S}}
 \toprule
Algorithm &   \multicolumn{1}{c}{40} &    \multicolumn{1}{c}{50}  &       \multicolumn{1}{c}{60}  &       \multicolumn{1}{c}{70}  &       \multicolumn{1}{c}{80}  &       \multicolumn{1}{c}{90}  &       \multicolumn{1}{c}{100} \\
  \midrule
AEO             & 5.933333 &  6.033333 &  6.033333 &  6.066667 &  6.133333 &  6.100000 &  6.100000 \\
EBOwithCMAR     & \bfseries 1.883333 & \bfseries 1.800000 & \bfseries 1.850000 & \bfseries 1.850000 & \bfseries 1.733333 & \bfseries 1.883333 & \bfseries 1.933333 \\
ELSHADE-SPACMA  & \bfseries 1.883333 &  2.166667 &  2.250000 &  2.366667 &  2.583333 &  2.650000 &  2.666667 \\
GSKA            & 4.000000 &  3.966667 &  4.066667 &  3.983333 &  3.933333 &  3.933333 &  3.883333 \\
PSO             & 5.766667 &  5.700000 &  5.566667 &  5.466667 &  5.466667 &  5.433333 &  5.366667 \\
SSA             & 5.600000 &  5.633333 &  5.633333 &  5.700000 &  5.666667 &  5.700000 &  5.733333 \\
jSO             & 2.933333 &  2.700000 &  2.600000 &  2.566667 &  2.483333 &  2.300000 &  2.316667 \\
  \bottomrule
  \end{tabular}
}

%% file: results_evol_cec_d30.tex
\subfloat[1-30\% evaluations]{
\begin{tabular}{l*{7}S}
\toprule
{} &       \multicolumn{1}{c}{1}   &       \multicolumn{1}{c}{2}   &       \multicolumn{1}{c}{3}   &       \multicolumn{1}{c}{5}   &       \multicolumn{1}{c}{10}  &       \multicolumn{1}{c}{20}  &       \multicolumn{1}{c}{30}\\
\midrule
AEO            &  2.333333 &  3.700000 &  4.533333 &  5.100000 &  5.566667 &  6.000000 &  6.100000\\
EBOwithCMAR    &  5.266667 &  4.666667 &  4.033333 &  3.466667 &  2.766667 & \bfseries 1.900000 &  2.066667\\
ELSHADE-SPACMA &  6.533333 &  5.633333 &  5.266667 &  4.166667 &  2.833333 &  1.933333 &  \bfseries 1.700000\\
GSKA           & \bfseries 1.633333 & \bfseries 1.400000 & \bfseries 1.400000 & \bfseries 1.433333 & \bfseries 2.100000 &  3.200000 &  3.966667\\
PSO            &  5.333333 &  6.066667 &  6.366667 &  6.500000 &  6.300000 &  6.166667 &  6.066667\\
SSA            &  2.300000 &  2.866667 &  3.200000 &  3.566667 &  4.566667 &  5.066667 &  5.200000\\
jSO            &  4.600000 &  3.666667 &  3.200000 &  3.766667 &  3.866667 &  3.733333 &  2.900000\\
\bottomrule
\end{tabular}
}
\\
\subfloat[40-100\% Evaluations]{
\begin{tabular}{l*{7}{S}}
\toprule
{} &       \multicolumn{1}{c}{40}  &       \multicolumn{1}{c}{50}  &       \multicolumn{1}{c}{60}  &       \multicolumn{1}{c}{70}  &       \multicolumn{1}{c}{80}  &       \multicolumn{1}{c}{90}  &       \multicolumn{1}{c}{100} \\
  \midrule
AEO             &  6.200000 &  6.266667 &  6.266667 &  6.266667 &  6.200000 &  6.166667 &  6.166667 \\
EBOwithCMAR     &  2.183333 &  2.366667 &  2.433333 &  2.316667 &  1.916667 &  1.916667 &  \bfseries 1.916667 \\
ELSHADE-SPACMA  & \bfseries 1.516667 & \bfseries 1.400000 & \bfseries 1.383333 & \bfseries 1.450000 & \bfseries 1.716667 & \bfseries 1.833333 &  2.133333 \\
GSKA            &  4.100000 &  4.200000 &  4.166667 &  4.166667 &  4.166667 &  4.166667 &  4.200000 \\
PSO             &  6.033333 &  5.966667 &  5.966667 &  5.966667 &  5.933333 &  5.933333 &  5.866667 \\
SSA             &  5.266667 &  5.400000 &  5.433333 &  5.466667 &  5.600000 &  5.633333 &  5.666667 \\
jSO             &  2.700000 &  2.400000 &  2.350000 &  2.366667 &  2.466667 &  2.350000 &  2.050000 \\
\bottomrule
\end{tabular}
}

%% file: results_evol_cec_d50.tex
\subfloat[1-30\% evaluations] {
\begin{tabular}{l*{7}{S}}
\toprule
{} &       \multicolumn{1}{c}{1}   &       \multicolumn{1}{c}{2}   &       \multicolumn{1}{c}{3}   &       \multicolumn{1}{c}{5}   &       \multicolumn{1}{c}{10}  &       \multicolumn{1}{c}{20}  &       \multicolumn{1}{c}{30}  \\
  \midrule
AEO            &  2.966667 &  4.200000 &  4.900000 &  5.466667 &  5.766667 &  5.966667 &  6.033333 \\
EBOwithCMAR    &  5.766667 &  4.566667 &  4.166667 &  3.666667 &  2.766667 &  2.266667 &  2.400000 \\
ELSHADE-SPACMA &  6.566667 &  5.900000 &  5.433333 &  4.366667 &  3.100000 &  \bfseries 2.066667 &  \bfseries 1.933333 \\
GSKA           & \bfseries 1.666667 & \bfseries 1.833333 & \bfseries 1.833333 & \bfseries 1.800000 & \bfseries 2.600000 &  3.700000 &  4.166667 \\
PSO            &  4.700000 &  5.900000 &  6.166667 &  6.400000 &  6.400000 &  6.233333 &  6.233333 \\
SSA            &  1.900000 &  2.166667 &  2.366667 &  2.733333 &  3.633333 &  4.400000 &  4.733333 \\
jSO            &  4.433333 &  3.433333 &  3.133333 &  3.566667 &  3.733333 &  3.366667 &  2.500000 \\
\bottomrule
\end{tabular}
}
\\
\subfloat[40-100\% evaluations]{
\begin{tabular}{l*{14}{S}}
\toprule
{} &       \multicolumn{1}{c}{40}  &       \multicolumn{1}{c}{50}  &       \multicolumn{1}{c}{60}  &       \multicolumn{1}{c}{70}  &       \multicolumn{1}{c}{80}  &       \multicolumn{1}{c}{90}  &       \multicolumn{1}{c}{100} \\
\midrule
AEO             &  6.066667 &  6.100000 &  6.100000 &  6.166667 &  6.233333 &  6.200000 &  6.166667 \\
EBOwithCMAR     &  2.466667 &  2.500000 &  2.566667 &  2.500000 &  2.233333 &  2.266667 &  2.166667 \\
ELSHADE-SPACMA  & \bfseries 1.766667 & \bfseries 1.366667 & \bfseries 1.233333 & \bfseries 1.200000 & \bfseries 1.300000 & \bfseries 1.383333 & \bfseries 1.783333 \\
GSKA            &  4.300000 &  4.333333 &  4.333333 &  4.400000 &  4.400000 &  4.400000 &  4.433333 \\
PSO             &  6.133333 &  6.100000 &  6.066667 &  6.033333 &  6.033333 &  6.000000 &  6.000000 \\
SSA             &  4.933333 &  5.033333 &  5.066667 &  5.133333 &  5.266667 &  5.333333 &  5.333333 \\
jSO             &  2.333333 &  2.566667 &  2.633333 &  2.566667 &  2.533333 &  2.416667 &  2.116667 \\
  \bottomrule
\end{tabular}
}

%% file: results_evol_cec_d100.tex
\subfloat[1-30\% evaluations] {
\begin{tabular}{l*{7}{S}}
\toprule
{} &       \multicolumn{1}{c}{1}   &       \multicolumn{1}{c}{2}   &       \multicolumn{1}{c}{3}   &       \multicolumn{1}{c}{5}   &       \multicolumn{1}{c}{10}  &       \multicolumn{1}{c}{20}  &       \multicolumn{1}{c}{30}  \\
  \midrule
AEO            &  3.400000 &  4.566667 &  5.133333 &  5.500000 &  5.733333 &  5.833333 &  5.800000 \\
EBOwithCMAR    &  5.833333 &  4.400000 &  4.100000 &  3.900000 &  3.100000 &  2.533333 &  2.700000 \\
ELSHADE-SPACMA &  6.400000 &  6.100000 &  5.333333 &  4.466667 &  3.700000 &  2.933333 &  2.433333 \\
GSKA           &  1.833333 &  \bfseries 1.933333 & \bfseries 1.966667 & \bfseries 2.166667 & \bfseries 2.766667 &  4.066667 &  4.133333 \\
PSO            &  4.433333 &  5.766667 &  6.300000 &  6.400000 &  6.400000 &  6.400000 &  6.433333 \\
SSA            &  \bfseries 1.633333 &  1.966667 &  2.066667 &  2.500000 &  2.966667 &  3.766667 &  4.166667 \\
jSO            &  4.466667 &  3.266667 &  3.100000 &  3.066667 &  3.333333 &  \bfseries 2.466667 &  \bfseries 2.333333 \\
\bottomrule
\end{tabular}
}
\\
\subfloat[40-100\% evaluations]{
\begin{tabular}{l*{14}{S}}
\toprule
{} &       \multicolumn{1}{c}{40}  &       \multicolumn{1}{c}{50}  &       \multicolumn{1}{c}{60}  &       \multicolumn{1}{c}{70}  &       \multicolumn{1}{c}{80}  &       \multicolumn{1}{c}{90}  &       \multicolumn{1}{c}{100} \\
\midrule
AEO             & 5.900000 &  5.966667 &  6.000000 &  6.033333 &  6.066667 &  6.066667 &  6.000000 \\
EBOwithCMAR     & 2.666667 &  2.500000 &  2.533333 &  2.433333 &  2.133333 &  2.233333 &  2.366667 \\
ELSHADE-SPACMA  & \bfseries 2.000000 &  \bfseries 1.433333 & \bfseries 1.333333 & \bfseries 1.400000 & \bfseries 1.466667 & \bfseries 1.433333 & \bfseries 1.383333 \\
GSKA            & 4.333333 &  4.433333 &  4.433333 &  4.466667 &  4.500000 &  4.500000 &  4.533333 \\
PSO             & 6.433333 &  6.400000 &  6.400000 &  6.400000 &  6.366667 &  6.366667 &  6.400000 \\
SSA             & 4.333333 &  4.533333 &  4.633333 &  4.633333 &  4.666667 &  4.733333 &  4.733333 \\
jSO             & 2.333333 &  2.733333 &  2.666667 &  2.633333 &  2.800000 &  2.666667 &  2.583333 \\
\bottomrule
\end{tabular}
}

%% file: comp_previous.tex
\begin{tabular}{lcccc}
\toprule
Func. &           Using SHADE         & Using  SaDE &   Using SHADE               &       IHDELS \\
&    +New Restart      &     +New Restart        &   +Old Restart    &                    \\
\midrule
$F_{1}$   &           2.69e-24 &           1.21e-24 &             1.76e-28 &  \textbf{4.80e-29} \\
$F_{2}$   &  \textbf{1.00e+03} &           1.26e+03 &             1.40e+03 &           1.27e+03 \\
$F_{3}$   &           2.01e+01 &           2.01e+01 &             2.01e+01 &  \textbf{2.00e+01} \\
\midrule
$F_{4}$   &  \textbf{1.48e+08} &           1.58e+08 &             2.99e+08 &           3.09e+08 \\
$F_{5}$   &  \textbf{1.39e+06} &           3.07e+06 &             1.76e+06 &           9.68e+06 \\
$F_{6}$   &  \textbf{1.02e+06} &           1.03e+06 &             1.03e+06 &           1.03e+06 \\
$F_{7}$   &  \textbf{7.41e+01} &           8.35e+01 &             2.44e+02 &           3.18e+04 \\
\midrule
$F_{8}$   &  \textbf{3.17e+11} &           3.59e+11 &             8.55e+11 &           1.36e+12 \\
$F_{9}$   &  \textbf{1.64e+08} &           2.48e+08 &             2.09e+08 &           7.12e+08 \\
$F_{10}$  &  \textbf{9.18e+07} &           9.19e+07 &             9.25e+07 &           9.19e+07 \\
$F_{11}$  &           5.11e+05 &  \textbf{4.76e+05} &             5.20e+05 &           9.87e+06 \\
\midrule
$F_{12}$  &  \textbf{6.18e+01} &           1.10e+02 &             3.42e+02 &           5.16e+02 \\
$F_{13}$  &  \textbf{1.00e+05} &           1.34e+05 &             9.61e+05 &           4.02e+06 \\
$F_{14}$  &  \textbf{5.76e+06} &           6.14e+06 &             7.40e+06 &           1.48e+07 \\
\midrule
  $F_{15}$  &  \textbf{6.25e+05} &           8.69e+05 &             1.01e+06 &           3.13e+06 \\
  \midrule
\textbf{Better} &  \multicolumn{1}{c}{12} & \multicolumn{1}{c}{1} & \multicolumn{1}{c}{0} & \multicolumn{1}{c}{2}\\
\bottomrule
\end{tabular}